\documentclass[preprint,12pt]{elsarticle}
\usepackage{lineno}
\usepackage{hyperref}
\modulolinenumbers[5]
\usepackage{amssymb}
\usepackage{amsmath}
\usepackage{epsfig}
\usepackage{psfrag}
\usepackage{graphics}
\usepackage{listings}
\usepackage{float}
\usepackage{tabularx}
\usepackage{caption}
\usepackage{subcaption}
\usepackage{amssymb,amsmath}
\usepackage{color}

\usepackage{multirow}
\usepackage{algorithm}
\usepackage[noend]{algpseudocode}
	\makeatletter
	\def\BState{\State\hskip-\ALG@thistlm}
	\makeatother
	
\usepackage{verbatim}
\usepackage[utf8]{inputenc}

\usepackage{tikz}
\usetikzlibrary{arrows}
\usetikzlibrary{shapes.geometric, arrows}
\usetikzlibrary{bayesnet}
\tikzstyle{startstop} = [rectangle, rounded corners, minimum width=3cm, minimum height=1cm,text centered, draw=black, fill=red!30]
\tikzstyle{io} = [trapezium, trapezium left angle=70, trapezium right angle=110, minimum width=3cm, minimum height=1cm, text centered, draw=black, fill=blue!30]
\tikzstyle{process} = [rectangle, minimum width=3cm, minimum height=1cm, text centered, draw=black, fill=orange!30]
\tikzstyle{decision} = [diamond, minimum width=3cm, minimum height=1cm, text centered, draw=black, fill=green!30]
\tikzstyle{arrow} = [thick,->,>=stealth]

\usetikzlibrary{mindmap,shadows}

\newdefinition{remark}{Remark}
\journal{Journal of Computational Physics}

\newcommand{\norm}[1]{\left| \left| {#1} \right| \right|}

\newcommand{\G}[3]{\mathcal N \left( {#1} | {#2}, {#3} \right)}

\newcommand{\GP}[2]{\mathcal{GP} \left( {#1}, {#2} \right)}

\newcommand{\KL}[2]{{\rm KL} \left({#1} \parallel {#2} \right)}
\newcommand{\mc}{\mathcal}
\newcommand{\mtx}{\mathbf}
\newcommand{\order}[1]{\mathcal O \left( {#1} \right)}
\newcommand{\reals}{\mathbb R}
\newcommand{\vc}{\boldsymbol}

\newcommand{\determinant}[1]{\left| {#1} \right|}

\newcommand{\diag}[0]{\textrm{diag}}

\newcommand{\expectation}[2]{\mathbb E_{#2} \left[ {#1} \right]}

\newcommand{\trace}[1]{\textrm{Tr}\left( {#1} \right)}

\newcommand{\xxi}[0]{\mtx X^{(\xi)}}
\newcommand{\xxis}[0]{\mtx X^{(\xi),*}}

\newcommand{\xsts}[0]{\mtx X^{(s),*}}
\newcommand{\xu}[0]{\mtx X_u}
\newcommand{\xuxi}[0]{\mtx X_u^{(\xi)}}
\newcommand{\xust}[0]{\mtx X_u^{(s)}}

\newcommand{\Ubar}[0]{\bar{\mtx U}}
\newcommand{\ubar}[0]{\bar{\vc u}}
\newcommand{\kff}[0]{\mtx K_{ff}}
\newcommand{\kyy}[0]{\mtx K_{yy}}
\newcommand{\kfu}[0]{\mtx K_{fu}}
\newcommand{\kuf}[0]{\mtx K_{uf}}
\newcommand{\kuu}[0]{\mtx K_{uu}}
\newcommand{\kuuinv}[0]{\kuu^{-1}}
\newcommand{\ksu}[0]{\mtx K_{*u}}
\newcommand{\kus}[0]{\mtx K_{u*}}
\newcommand{\ksf}[0]{\mtx K_{*f}}
\newcommand{\kfs}[0]{\mtx K_{f*}}
\newcommand{\kss}[0]{\mtx K_{**}}
\newcommand{\ktilde}[0]{\tilde{\mtx K}}

\newcommand{\psione}[0]{\mtx \Psi_1}
\newcommand{\psitwo}[0]{\mtx \Psi_2}
\newcommand{\kpsi}[0]{\mtx K_\psi}
\newcommand{\kpsiinv}[0]{\kpsi^{-1}}

\begin{document}

\begin{frontmatter}

\title{Structured Bayesian Gaussian process latent variable model: applications to data-driven dimensionality reduction and high-dimensional inversion}
\author[ad1]{Steven Atkinson}
\ead{satkinso@nd.edu}
\author[ad1]{Nicholas Zabaras\corref{cor}}
\ead{nzabaras@gmail.com}
\ead[url]{https://cics.nd.edu/}
\cortext[cor]{Corresponding author}
\address[ad1]{
    Center for Informatics and Computational Science \\ 
    University of Notre Dame \\
    311 I Cushing Hall, 
    Notre Dame, IN 46556, USA
}
\date{\today}

\begin{abstract}
We introduce a methodology for nonlinear inverse problems using a variational Bayesian approach where the unknown quantity is a spatial field.
A structured Bayesian Gaussian process latent variable model is used both to construct a low-dimensional generative model of the sample-based stochastic prior as well as a surrogate for the forward evaluation.
Its Bayesian formulation captures epistemic uncertainty introduced by the limited number of input and output examples, automatically selects an appropriate dimensionality for the learned latent representation of the data, and rigorously propagates the uncertainty of the data-driven dimensionality reduction of the stochastic space through the forward model surrogate.
The structured Gaussian process model explicitly leverages spatial information for an informative generative prior to improve sample efficiency while achieving computational tractability through Kronecker product decompositions of the relevant kernel matrices.
Importantly, the Bayesian inversion is carried out by solving a variational optimization problem, replacing traditional computationally-expensive Monte Carlo sampling.
The methodology is demonstrated on an elliptic PDE and is shown to return well-calibrated posteriors and is tractable with latent spaces with over $100$ dimensions.
\end{abstract}


\end{frontmatter}

\section{Introduction}
\label{sec:Intro}

Inverse problems are ubiquitous in computational modeling and arise when one is interested in determining some quantity of interest, but only has access to indirect measurements that are the outcome of some physical process.
Examples of inverse problems include parameter calibration~\cite{kennedy2001bayesian, kaipio2006statistical}, 
subsurface flow~\cite{berger2003markov, aarnes2007modelling},
ocean dynamics~\cite{mckeague2005statistical},
remote sensing~\cite{haario2004markov},
seismic inversion~\cite{martin2012stochastic},
and others.
The Bayesian statistical formulation has been fruitful for providing a means of rigorously approaching inverse problems.
One defines a prior probability measure over the quantity of interest, then seeks to determine the posterior effected by the observations data available.

An analytical treatment for Bayesian inversion is not possible in most problems of interest owing to the complexity of the system.
In the basic Monte Carlo approach to Bayesian inversion, one aims to generate samples from the posterior distribution over the stochastic parameters by using the unnormalized posterior, i.e., the product of the prior and likelihood).
The prior is usually defined in some mathematically-convenient way, and the likelihood is computed by evaluating a forward model (e.g.\ solving a system of differential equations).
This, in principle, provides a straightforward means of solving inverse problems.

There are two practical challenges to the approach described above.
First, the forward model typically models a nontrivial physical system and therefore is usually costly to evaluate from a computational standpoint.
Thus, for computational tractability, one may construct a cheap surrogate model that incurs some (hopefully controlled) loss in accuracy in exchange for a substantial reduction in computational cost.
Surrogate models such as polynomial chaos~\cite{marzouk2007stochastic, marzouk2009dimensionality}, proper orthogonal decomposition~\cite{cui2015data}, or Gaussian processes~\cite{bilionis2013multi, bilionis2013solution} have been studied.

The second practical challenge is that the stochastic space may be high-dimensional.
In the case where one seeks to infer an unknown field, the infinite-dimensional function space is customarily projected onto finite-dimensional space through the discretization of the problem on a finite element mesh;
Consequently, one then must infer the posterior in a space with dimensionality corresponding to the discretization.
Directly exploring this high-dimensional space is computationally intractable in practice.
To overcome this, one might define a dimensionality reduction scheme that projects the original stochastic space into a lower-dimensional space exploiting correlations in the stochastic variables' prior.
One notable example of this is to use a Karhun\`en-Loeve expansion (KLE)~\cite{lord2014introduction} to parameterize the prior with relatively few latent variables; this has been explored in the context of Bayesian inversion in~\cite{marzouk2009dimensionality}.

However, this choice is restrictive on the possible choices of prior (e.g.\ a Gaussian process with some predetermined mean function and kernel) and may not accurately reflect one's true prior beliefs and (lack of) knowledge.
For example, in the permeability estimation problem, it is known that the unknown field may have nontrivial correlation statistics~\cite{winter2000mean, winter2002groundwater, winter2003moment, winter2006multivariate}.
To combat this, a wavelet-based method that allows for modeling multiple length scales was introduced~\cite{ellam2016bayesian} that provided a general approach to solving the so-called ``scale determination'' problem.
Still, such a representation is based on mathematical convenience and may need to be overly broad to represent a more compact subspace on which feasible inputs may be known to lie.
Striking a balance between a suitably broad prior so as not to rule out the ground truth to be inferred while simultaneously leveraging physical knowledge to exclude implausible regions of the input space is the key to data-efficient Bayesian inversion in high dimensions.

In the current work, we consider the case in which one may have access to some physical model that can be used to generate realistic samples from the input density implicitly.
To do this, we utilize a structured Gaussian process latent variable model (SGPLVM)~\cite{atkinson2018structured}.
The SGPLVM is a Bayesian, nonlinear, nonparametric model that can be used to learn the underlying structure in a set of observations.
Moreover, the SGPLVM explicitly models spatial correlations in the observation data through parametric kernel functions, not only improving its sample efficiency but yielding an interpretable learned representation.
Its Bayesian extension~\cite{titsias2010bayesian} additionally captures the epistemic uncertainty about the learned latent variables within a variational framework by defining a variational posterior distribution over the latent variables and deriving a lower bound to the model evidence which allows one to approximately integrate over the uncertain inputs.

We note that neural networks have made great progress as generative models in recent years via variational autoencoder architectures~\cite{kingma2013auto} and adversarial formulations~\cite{goodfellow2014generative}. 
A main focus of this work is to sample efficiency within unsupervised learning;
as of now, the above models and derived extensions generally struggle to perform well with a limited supply of data.
A reasonable means of addressing this challenge is by carefully-chosen Bayesian regularization;
however, we are still challenged to pose tractable, interpretable priors and methods for inference.
While imposing Gaussian or Laplace priors over the model weights~\cite{blundell2015weight} is attractive because of its mathematical tractability~\cite{goodfellow2016deep}, one may still question 
how well one understands the distributions over functions they imply~\cite{salimbeni2017doubly}.
Indeed, merely \textit{interpreting} tractable neural network priors is a challenging task that is of considerable interest in its own right.
By contrast, our approach in this work, based in Gaussian process models, begins by defining the generative model's distribution over functions and proceeding to derive a tractable means of inference.

The main contributions of this work are as follows.
First, we derive a structured Gaussian process latent variable model (SGPLVM) extending the work of Titsias and Lawrence~\cite{titsias2010bayesian} by explicitly accounting for spatial correlations in observations through parameterized kernels.
Computational tractability is maintained by exploiting the structure of the covariance matrix by extending the structured GP methods first introduced by Saat\c{c}i~\cite{saatcci2012scalable}.
We then show how one can use a pair of SGPLVM ``submodels'' to model an elliptic stochastic PDE in which one model performs data-driven dimensionality reduction on a provided high-dimensional conductivity field and the other learns a regression, given the learned latent variables, to the solution of the PDE.
Finally, we show how inverse problems using the same system can be efficiently solved (i.e.\ inferring an unknown conductivity field, given noisy observations of the solution to the PDE at a few spatial points).
Importantly, the variational formulation allows us to infer a variational posterior latent variable for the observation based on the incomplete noisy output data provided and propagate the uncertainty it implies in a rigorous way to the resulting posterior over the high-dimensional physical input space.

The rest of this paper is organized as follows.
In Section~\ref{sec:Theory}, we define the model structure of the structured GP-LVM, derive a variational lower bound that may be maximized to train the model, describe how one may use the trained model to make predictions, and explain how the developed model may be applied to the specific case of data-driven modeling for the solution of Bayesian inverse problems.
In Section~\ref{sec:Examples}, we apply our model to the task of solving both the forward and inverse problems on an elliptic PDE for which the high-dimensional conductivity is unknown and its prior distribution is only implicitly known through a generative process.
Section~\ref{sec:Conclusion} summarizes our work and discusses its implications.

\section{Theory of Gaussian Process Regression}
\label{sec:Theory}

In this section, we derive the structured GP-LVM (SGP-LVM) model.
For the sake of completeness, we first review GP regression in Section~\ref{sec:Theory:GPR} and the structured GP regression model in Section~\ref{sec:Theory:SGPR}.
Then, we present the novel structured Bayesian Gaussian process latent variable model in Section \ref{sec:Theory:SGPLVM}.
Finally, we define the inverse problem in Section~\ref{sec:Theory:Inverse} and explain how the SGPLVM can be used to solve it.

Before proceeding, we quickly define some useful notation.
Let $\mtx X$ be a matrix with $i$-th row $\vc x_{i, :}$, $j$-th column $\vc x_{:, j}$, and entry $x_{ij}$.
For compactness of notation, we may index a set of vectors with a single index, e.g., $\vc x_i$, when the distinction between row and column vectors is unimportant.
The matrix $\mtx X$ may be assembled by a set of vectors $\vc x$ as rows of $\mtx X$.

\subsection{Gaussian process regression}
\label{sec:Theory:GPR}

A Gaussian process is a distribution over functions 
$f(\cdot) = \GP {\mu(\cdot)} {k(\cdot, \cdot; \vc \theta_x)}: \mc X \rightarrow \reals$
defined by a mean function $\mu$ and a kernel function $k$ with hyperparameters $\vc \theta_\mu$ and $\vc \theta_k$, respectively.
We are interested in modeling the unknown function 
\begin{equation}
    y(\vc x) = f(\vc x) + \epsilon,
\end{equation}
where $\epsilon \sim \G{\epsilon}{0}{\beta^{-1}}$ models corrupting noise with a spherical Gaussian distribution.
We refer to $f$ as the latent output and distinguish it from the observed output, which may be corrupted by noise.
A trivial extension of this model to multivariate outputs $\vc y \in \reals^{d_y}$
is \begin{equation}
    y_{:, j}(\vc x) = f_{:, j}(\vc x) + \epsilon, ~ j=1, \dots, d_y,
    \label{eqn:Theory:GPR:Process}
\end{equation}
$f_{:, j}(\cdot) = \GP{\vc 0}{k(\cdot, \cdot; \vc \theta_x)}: \mc X \rightarrow \reals$ for $j=1, \dots, d_y$.
In other words, the various output dimensions share the same kernel, but are otherwise uncorrelated.

Given some inputs $\mtx X \in \reals^{n \times d_x}$ and corresponding outputs $\mtx Y \in \reals^{n \times d_y}$, the likelihood is
\begin{equation}
    \begin{aligned}
        p(\mtx Y | \mtx X, \vc \theta) &= \prod_{j=1}^{d_y} \G{\vc y_{:, j}}{\vc 0}{\mtx K_{yy}},
        \\
        \mtx K_{yy} &= \mtx K_{ff} + \beta^{-1} \mtx I_{n \times n},
        \\
        (\mtx K_{ff})_{ij} &= k(\vc x_{i, :}, \vc x_{j, :}; \vc \theta_x).
    \end{aligned}
    \label{eqn:Theory:GPR:Likelihood}
\end{equation}
One trains the model by optimizing the log-likelihood over the model hyperparameters $\vc \theta = \{\vc \theta_\mu, \ \theta_k, \beta \}$.
Notice that the GP regression model captures correlations between data (rows of $\mtx Y$), but not between dimensions (columns of $\mtx Y$).

The posterior predictive density evaluated at some test points $\mtx X^{n^* \times d_x}$ is
\begin{align}
    p(\mtx Y^* | \mtx X^*, \mtx X, \mtx Y, \vc \theta) &= \prod_{j=1}^{d_y} \G{\vc y_{:, j}^*}{\vc m_{:, j}^*}{\mtx C^*},
    \\
    \mtx M^* &= \mtx K_{*f} \mtx K_{yy}^{-1} \mtx Y,
    \\
    \mtx C^* &= \mtx K_{**} - \mtx K_{*f} \mtx K_{yy}^{-1} \mtx K_{f*} + \beta^{-1} \mtx I_{n^* \times n^*},
    \\
    (\mtx K_{f*})_{ij} &= k(\vc x_{i, :}, \vc x_{j, :}^*; \vc \theta_k),
    \label{eqn:Theory:GPR:Predictions}
\end{align}
and $\mtx K_{*f} = \mtx K_{f*}^\intercal$.
For more information, the interested reader may refer to~\cite{rasmussen2006gaussian}.

\subsection{Structured Gaussian process regression}
\label{sec:Theory:SGPR}

The structured Gaussian process regression (SGPR) model~\cite{saatcci2012scalable} begins by assuming that the training input locations are expressed as a Cartesian product.\footnote{In the original formulation~\cite{saatcci2012scalable}, it was assumed that each component in the Cartesian product be one-dimensional; however, this may be trivially extended to multidimensional components as was done in~\cite{bilionis2013multi}.}
In anticipation of our application, we define
\begin{equation}
    \mtx X = \mtx X^{(\xi)} \times \mtx X^{(s)},
\end{equation}
where $\mtx X^{(\xi)} \in \reals^{n_\xi \times d_\xi}$ contains latent variable inputs\footnote{In the context of this work, where we will be modeling a set of runs of a computer code, the latent variables serve to identify the parameterization of each run.}, $\mtx X^{(s)} \in \reals^{n_s \times d_s}$ contains the spatial points on which the outputs are observed, and the $\times$ operation denotes the Cartesian product of the rows of the two matrices provided as operands.
Note that we might further decompose $\mtx X^{(s)}$ along each spatial dimension:
\begin{equation}
    \mtx X^{(s)} = \mtx X^{(s, 1)} \times \dots \times \mtx X^{(s, d_s)},
\end{equation}
provided that the spatial points exhibit such structure.
However, we will not consider this during our derivations for the sake of conciseness.

The second assumption in building the SGPR model is that the kernel be separable:
\begin{equation}
    k(\vc x_{i, :}, \vc x_{j, :}) = \prod_{l \in \{\xi, s\}} k_l (\vc x_{i_l, :}^{(l)}, \vc x_{j_l, :}^{(l)}; \vc \theta_{k_l}),
    \label{eqn:SeparableKernel}
\end{equation}
where the indices $i_l$ and $j_l$ select the appropriate rows from $\mtx X^{(l)}$.

\subsection{Training}
Under these assumptions, the model likelihood is still given by Eq.~(\ref{eqn:Theory:GPR:Likelihood}).
However, the covariance matrix for the latent outputs, $\kff$, can now be written as a Kronecker product~\cite{saatcci2012scalable, bilionis2013multi}:
\begin{equation}
    \kff = \kff^{(\xi)} \otimes \kff^{(s)}.
\end{equation}
Furthermore, we know from the properties of Kronecker products \cite{brewer1978kronecker} that its eigendecomposition is given as
\begin{align}
    \kff &= \mtx Q \mtx \Lambda \mtx Q^\intercal
    \\
    &= \left( \mtx Q^{(\xi)} \otimes \mtx Q^{(s)} \right)
    \left( \mtx \Lambda^{(\xi)} \otimes \mtx \Lambda^{(s)} \right)
    \left( \mtx Q^{(\xi)\intercal} \otimes \mtx Q^{(s)\intercal} \right),
\end{align}
where the columns of the orthogonal matrices $\mtx Q$, $\mtx Q^{(\xi)}$, and $\mtx Q^{(s)}$ are the eigenvectors of $\kff$, $\kff^{(\xi)}$, and $\kff^{(s)}$, respectively; the eigenvalues are given by the respective entries of the diagonal matrices $\mtx \Lambda$, $\mtx \Lambda^{(\xi)}$, and $\mtx \Lambda^{(s)}$.
The model covariance matrix may then be written as
\begin{align}
    \kyy &= \kff + \beta^{-1} \mtx I_{n \times n}
    \\
    &= \mtx Q (\mtx \Lambda + \beta^{-1} \mtx I_{n \times n}) \mtx Q^\intercal,
\end{align}
and its inverse can be expressed as
\begin{equation}
    \kyy^{-1} = \mtx Q (\mtx \Lambda + \beta^{-1} \mtx I_{n \times n})^{-1} \mtx Q^\intercal.
    \label{eqn:SGPR:KyyInverse}
\end{equation}
Note that the matrix $\mtx \Lambda + \beta^{-1} \mtx I_{n \times n}$ is diagonal and can therefore be easily inverted on an element-wise basis.
Furthermore, the log-determinant is given as simply
\begin{align}
    \log |\kyy| = \sum_{i=1}^n \log(\lambda_{ii} + \beta^{-1}).
    \label{eqn:Theory:SGPR:LogKyy}
\end{align}
Given these, we see that the model likelihood of Eq.\ (\ref{eqn:Theory:GPR:Likelihood}) can be evaluated for the SGPR model with a computational cost that scales linearly with $n d_y$ in time and memory by utilizing Eq.\ (\ref{eqn:Theory:SGPR:LogKyy}) and exploiting the Kronecker structure of $\mtx Q$ along with fast Kronecker matrix-matrix products~\cite{saatcci2012scalable}.

\subsection{Predictions with SGPR}
Consider $n^*$ test inputs formed by a Cartesian product of $n_\xi^*$ stochastic test points and $n_s^*$ spatial test points:
\begin{equation}
    \mtx X^* = \mtx X^{(\xi, *)} \times \mtx X^{(s, *)}.
    \label{eqn:SGPR:TestInputs}
\end{equation}
The predictive density is the same as in Eq.~(\ref{eqn:Theory:GPR:Predictions}).
However, due to the structure assumed by Eq.~(\ref{eqn:SGPR:TestInputs}), the test kernel matrices possess Kronecker product structure:
\begin{align}
    \mtx K_{*f} &= \mtx K_{*f}^{(\xi)} \otimes \mtx K_{*f}^{(s)},
    \\
    \mtx K_{**} &= \mtx K_{**}^{(\xi)} \otimes \mtx K_{**}^{(s)},
\end{align}
The predictive mean in Eq.~(\ref{eqn:Theory:GPR:Predictions}) can be computed in $\order{n n^*}$ time and memory using the Kronecker matrix-matrix product algorithm provided in~\cite{saatcci2012scalable}.

If we are interested only in the marginals over the test outputs, then the predictive variance $\textrm{diag}(\mtx \Sigma^*) \in \reals^{n^* \times 1}$ may be computed efficiently as well:
\begin{equation}
    \textrm{diag}(\mtx \Sigma^*) = \textrm{diag}(\mtx K_{**}) - \left( (\mtx K_{*f} \mtx Q) \circ (\mtx K_{*f} \mtx Q) \right) \textrm{diag}\left( \kyy^{-1} \right),
    \label{eqn:Theory:SGPR:PredictiveVariance}
\end{equation}
where $\mtx A \circ \mtx B$ denotes the Schur product between $\mtx A$ and $\mtx B$; by noticing that
\begin{equation}
    (\mtx K_{*f} \mtx Q) \circ (\mtx K_{*f} \mtx Q) = 
    \left( (\mtx K_{*f}^{(\xi)} \mtx Q^{(\xi)}) \circ (\mtx K_{*f}^{(\xi)} \mtx Q^{(\xi)}) \right)
    \otimes
    \left( (\mtx K_{*f}^{(s)} \mtx Q^{(s)}) \circ (\mtx K_{*f}^{(s)} \mtx Q^{(s)}) \right),
\end{equation}
\noindent the full predictive covariance can be computed in $\order{(n^*)^2 n}$ time and $\order{(n^*)^2}$ memory; see~\ref{apx:SGPR:PredictiveCovariance} for details.

\noindent
\begin{remark}
    The linear scaling of the computational cost associated with training the SGPR model assumes that $n_\xi$ and $n_s$ are sufficiently small that the main computational bottleneck is evaluating the matrix product $\mtx Q^\intercal \mtx Y$ found by substituting Eq.~(\ref{eqn:SGPR:KyyInverse}) into Eq.~(\ref{eqn:Theory:GPR:Likelihood}).
    This operation has a computational cost that scales as $\order{n d_y}$ in time and memory.
    This assumption may be violated if either $n_\xi$ or $n_s$ become too large, in which case the bottleneck shifts to that associated with the eigendecompositions of $\kff^{(\xi)}$ and/or $\kff^{(s)}$, which scale as $\order{n_\xi^3}$ and $\order{n_s^3}$, respectively.
    If one can also find Toeplitz structure in the spatiotemporal points, then this can be further prolonged~\cite{cunningham2008fast, turner2010statistical, wilson2015kernel}, but no such structure will exist in general for the high-dimensional stochastic inputs in $\mtx X^{(\xi)}$.
    We note that Toeplitz structure-exploiting techniques might be applied on top of the methods used in this paper from a theoretical perspective, but we do not consider them here.
\end{remark}

\noindent
\begin{remark}
    The structured GP model is mathematically quite similar to a multi-output Gaussian process model using the linear model of coregionalization (LMC)~\cite{journel1978mining, goovaerts1997geostatistics}.
    In the LMC model, a coregionalization matrix captures the correlations between output dimensions that we have here reinterpreted as many \textit{data}.
    An important distinction is that we are able to define a kernel over the spatiotemporal locations that reflects our physical intuition that our realizations will exhibit spatiotemporal correlations that will decay in a particular manner.
    By parameterizing these correlations with a kernel function, we greatly reduce the number of parameters that must be trained, as well as illuminate the possibility that the structure may be further decomposed along individual spatiotemporal dimensions, provided that the input data exhibits such structure.
    For example, na\"ive use of the LMC model on observations over a two-dimensional $64 \times 64$ lattice would result in needing to learn $(64^2 (64^2 - 1))/2 \approx 8 \times 10^6$ parameters for a positive-definite coregionalization matrix; by parameterizing the spatial correlations instead with a separable, stationary kernel, we instead only have to learn $2$ parameters (the length scales for the two input dimensions).
    This simplification is essential to obtaining a model for which training is computationally tractable for the problems we are interested in.
    Furthermore, the use of a parameterized kernel enables us to predict at spatial locations that are not included in the training data.
\end{remark}

\section{Structured Bayesian Gaussian process latent variable model}
\label{sec:Theory:SGPLVM}

In this subsection, we introduce the SGPLVM.
The theoretical discussion of the model is broken into a description of the model architecture and derivation of the evidence lower bound in Section~\ref{sec:Theory:SGPLVM:Training}, 
its efficient computation using structure-exploiting algebra in Section~\ref{sec:Theory:SGPLVM:TrainingKronecker}, and its subsequent use for making predictions in 
Section~\ref{sec:Theory:SGPLVM:Predictions}.

\subsection{SGPLVM: model architecture and evidence lower bound}
\label{sec:Theory:SGPLVM:Training}

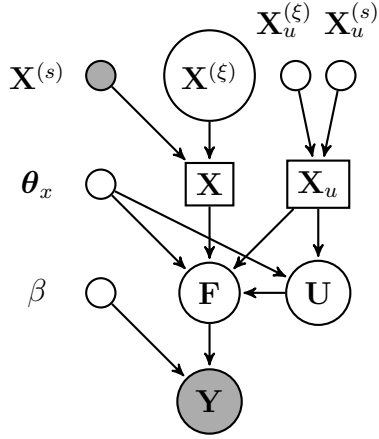
\begin{figure}
    \centering
    \begin{tikzpicture}[->,
        >=stealth',
        shorten >=1pt,
        auto,
        node distance=3.5em, 
        thick,
        observed/.style={
            circle,
            minimum size=0.05cm,
            draw,
            fill=gray!65
        },
        unobserved/.style={
            circle,
            minimum size=0.05cm,
            draw
        },
        deterministic/.style={
            rectangle,
            draw
        }
        ] 
        \node[unobserved]    (x_xi) {$\xxi$};
        \node[observed]      (x_st) [left of=x_xi] {};
        \node                (x_st_label) [left of=x_st, xshift=0.6cm] {$\mtx X^{(s)}$};
        \node[deterministic] (x) [below of=x_xi]  {$\mtx X$};
        \node[unobserved]    (theta_x) [left of=x] {};
        \node                (theta_x_label) [left of=theta_x, xshift=0.6cm] {$\vc \theta_x$};
        \node[deterministic] (xu) [right of=x] {$\mtx X_u$};
        \node[unobserved]    (xu_xi) [above of=xu, xshift=-0.3cm] {};
        \node                (xu_xi_label) [above of=xu_xi, xshift=-0.15cm, yshift=-0.7cm] {$\xuxi$};
        \node[unobserved]    (xu_st) [above of=xu, xshift=0.3cm] {};
        \node                (xu_st_label) [above of=xu_st, xshift=0.15cm, yshift=-0.7cm] {$\xust$};
        \node[unobserved]    (f) [below of=x] {$\mtx F$};
        \node[unobserved]    (fu) [right of=f] {$\mtx U$};
        \node[unobserved]    (beta) [left of=f] {};
        \node                (beta_label) [left of=beta, xshift=0.6cm] {$\beta$};
        \node[observed]      (y) [below of=f] {$\mtx Y$};
    
        \path
            (x_xi) edge (x)
            (x_st) edge (x)
            (xu_xi) edge (xu)
            (xu_st) edge (xu)
            (x) edge (f)
            (xu) edge (fu)
            (theta_x) edge (f)
            (theta_x) edge (fu)
            (xu) edge (f)
            (fu) edge (f)
            (f) edge (y)
            (beta) edge (y)
            ;
    \end{tikzpicture}
    \caption{
        Probabilistic graphical model for the SGPLVM.
        Observed variables are denoted with shaded nodes, while unobserved variables are shown as white nodes.
        Note that $\mtx X$ contains both observed dimensions (corresponding to spatiotemporal locations) and unobserved dimensions (corresponding to the latent variables).
        Deterministic operations (e.g.\ Cartesian products) are shown as boxes.
        Small nodes denote variables that are modeled as points, and large nodes denote variables modeled with distributions.
    }
    \label{fig:Theory:SGPLVM:PGM}
\end{figure}

The SGPLVM is a generative model that differs from the SGPR in that we assume that data are provided as unlabeled observations; our goal is to infer some inputs that, when propagated through the structured GP as in Eq.~(\ref{eqn:Theory:GPR:Process}), generate the observations as output.
Thus, our inputs are again comprised of latent variables $\mtx X^{(\xi)}$ along with spatial points $\mtx X^{(s)}$ at which the data are observed.
The spatial inputs are known exactly (e.g.\ as coordinates of pixels in an image); however, one must infer what the latent variables $\mtx X^{(\xi)}$ ought to be.
Thus, we regard them as uncertain and assign to them a spherical Gaussian prior~\cite{titsias2010bayesian}:
\begin{equation}
    p(\xxi) = \prod_{i=1}^{n_\xi} \prod_{j=1}^{d_\xi} \G{x_{ij}^{(\xi)}}{0}{1}.
    \label{eqn:Theory:SGPLVM:pxi}
\end{equation}
In order to train the model, we seek to evaluate the model evidence
\begin{equation}
    p(\mtx Y) = \int p(\mtx Y | \mtx X) p(\xxi) d \mtx \xxi.
\end{equation}
However, integrating over $\xxi$ is intractable.
Therefore, the model is augmented with $m$ inducing input-output pairs, assembled in matrices $\mtx X_u \in \reals^{m \times d_x}$ and $\mtx U \in \reals^{m \times d_y}$.
We assume that these inducing pairs are modeled by the same generative process as the training inputs and latent outputs $\mtx F \in \reals^{n \times d_y}$, allowing us to write the following joint probability:
\begin{equation}
    p(\mtx F^+ | \mtx X^+) = \prod_{j=1}^{d_y} \G{\vc f_{:, j}^+}{\vc m_{:, j}^+}
    {\left( \begin{matrix} \kff & \kfu \\ \kuf & \kuu \end{matrix} \right)},
\end{equation}
where
\begin{equation}
    \begin{aligned}
        \mtx X^+ &= \left( \begin{matrix} \mtx X \\ \mtx X_u \end{matrix} \right),
        \\
        \mtx F^+ &= \left( \begin{matrix} \mtx F \\ \mtx U \end{matrix} \right),
    \end{aligned}
\end{equation}
$\mtx M^+$ similarly concatenates the mean function evaluated over the training inputs and inducing inputs, and $\kuf$, $\kfu$, and $\kuu$ are likewise evaluated using the kernel over the training inputs and inducing inputs as expected.
We also note the conditional GP prior
\begin{equation}
    p(\mtx F | \mtx X, \mtx U, \mtx X_u) = \prod_{j=1}^{d_y} \G{\vc f_{:, j}}{\vc \eta_{:, j}}{\ktilde},
    \label{eqn:Theory:SGPLVM:ConditionalGPPrior}
\end{equation}
where the conditional mean and covariance take the usual forms from the projected process model \cite{seeger2003fast, rasmussen2006gaussian}:
\begin{align}
    \vc \eta &= \kfu \kuuinv \mtx Y,
    \\
    \ktilde &= \kff - \kfu \kuuinv \kuf.
\end{align}
Finally, we define the Gaussian variational posterior over the induced outputs
\begin{equation}
    q(\mtx U) = \prod_{j=1}^{d_y} \G{\vc u_{:, j}}{\ubar_{:, j}}{\mtx \Sigma_u},
\end{equation}
and pick a joint variational posterior that factorizes as
\begin{equation}
    q(\mtx F, \mtx U, \xxi) = q(\mtx F | \xxi, \mtx U) q(\mtx U) q(\xxi).
\end{equation}
By assuming that the inducing points are sufficient statistics of the training data, then we can let $q(\mtx F | \xxi, \mtx U)$ take the form of the conditional GP prior of Eq.~(\ref{eqn:Theory:SGPLVM:ConditionalGPPrior}).
Finally, we pick the following variational posterior for the latent variables:
\begin{equation}
    q(\xxi) = \prod_{i=1}^{n_\xi} \prod_{j=1}^{d_\xi} \G{x_{ij}^{(\xi)}}{\mu_{ij}^{(\xi)}}{s_{ij}^{(\xi)}}.
    \label{eqn:Theory:SGPLVM:qxi}
\end{equation}
A probabilistic graphical model for the SGPLVM is shown in Fig.~\ref{fig:Theory:SGPLVM:PGM}.
Note that one may also model dynamical data by introducing a GP prior on $\xxi$ if the observations are associated with known time points; see~\cite{atkinson2018fully}.

Using the familiar variational approach~\cite{bishop2006pattern, titsias2010bayesian}, we use Jensen's inequality to write a lower bound for the logarithm of the model evidence:
\begin{equation}
    \log p(\mtx Y) \ge \mc L = \int q(\mtx F, \mtx U, \xxi) \log \frac{p(\mtx Y, \mtx F, \mtx U, \xxi)}{q(\mtx F, \mtx U, \xxi)} d \mtx F d \mtx U d \xxi.
\end{equation}
After some manipulations, one arrives at the following uncollapsed lower bound:
\begin{equation}
    \begin{aligned}
        \mc L = &-\frac{n d_y}{2} \left( \log(2 \pi) - \log \beta \right)
        -\frac{\beta}{2} \trace{\mtx Y \mtx Y^\intercal}
        + \beta \trace{\Ubar^\intercal \kuuinv \psione^\intercal \mtx Y}
        \\
        &
        - \frac{\beta d_y}{2} \trace{\kuuinv \psitwo \kuuinv \left( \Ubar \Ubar^\intercal + d_y \mtx \Sigma_u \right)}
        - \frac{\beta d_y}{2} \left( \psi_0 - \trace{\kuu^{-1} \psitwo} \right)
        \\
        & - \KL{q(\mtx U)}{p(\mtx U)} - \KL{q(\xxi)}{p(\xxi)},
    \end{aligned}
    \label{eqn:Theory:SGPLVM:UncollapsedBound}
\end{equation}
where we have defined the kernel expectations
\begin{equation}
    \begin{aligned}
        \psi_0 &= \expectation{\trace{\kff}}{q(\xxi)},
        \\
        \psione &= \expectation{\kfu}{q(\xxi)},
        \\
        \psitwo &= \expectation{\kuf \kfu}{q(\xxi)}.
    \end{aligned}
    \label{eqn:Theory:SGPLVM:PsiStatistics}
\end{equation}
Using standard methods~\cite{bishop2006pattern}, one may find an analytic optimum for $q(\mtx U)$ that maximizes Eq.~(\ref{eqn:Theory:SGPLVM:UncollapsedBound}) with respect to the variational parameters $\Ubar$ and $\mtx \Sigma_u$:
\begin{align}
    q^*(\mtx U) &= \prod_{j=1}^{d_y} \G{\vc u_{:, j}}{\ubar_{:, j}^*}{\mtx \Sigma_u^*},
    \label{eqn:Theory:SGPLVM:OptimalU}
    \\
    \Ubar^* &= \kuu \kpsiinv \psione^\intercal \mtx Y,
    \label{eqn:Theory:SGPLVM:OptimalU:Mean}
    \\
    \mtx \Sigma^* &= \beta^{-1} \kuu \kpsiinv \kuu,
    \label{eqn:Theory:SGPLVM:OptimalU:Covariance}
\end{align}
where we have defined
\begin{equation}
    \kpsi =  \beta^{-1} \kuu + \psitwo,
    \label{eqn:Theory:SGPLVM:KPsi}
\end{equation}
for convenience.
Substituting this into Eq.~(\ref{eqn:Theory:SGPLVM:UncollapsedBound}) and doing the required manipulations results in the ``collapsed'' lower bound:
\begin{equation}
    \begin{aligned}
        \mc L = &\frac{d_y}{2} \left(
            (n - m) \log \beta - n \log (2 \pi) - \log \determinant{\mtx A}
        \right)
        \\
        &- \frac{\beta}{2} \left(
            \trace{\mtx Y \mtx Y^\intercal}
            - \trace{\mtx Y^\intercal \mtx \Psi_1 \mtx K_\psi^{-1} \mtx \Psi_1^\intercal \mtx Y}
            + d_y \left(
                \psi_0 - \trace{\mtx C}
            \right)
        \right)
        \\
        &- \KL{q(\xxi)}{p(\xxi)},
    \end{aligned}
    \label{eqn:Theory:SGPLVM:CollapsedBound}
\end{equation}
where 
\begin{align}
    \mtx C &= \mtx L^{-1} \psitwo \mtx L^{-\intercal},
    \label{eqn:Theory:SGPLVM:C}
    \\
    \mtx A &= \mtx L^{-1} \kpsi \mtx L^{-\intercal} = \beta^{-1} \mtx I_{m \times m} + \mtx C,
    \label{eqn:Theory:SGPLVM:A}
\end{align}
and 
\begin{equation}
    \kuu = \mtx L \mtx L^\intercal,  
    \label{eqn:Theory:SGPLVM:L}
\end{equation}
is a Cholesky decomposition.
Given the forms specified in Eqs.~(\ref{eqn:Theory:SGPLVM:pxi}) and~(\ref{eqn:Theory:SGPLVM:qxi}), the KL divergence is given as
\begin{equation}
    \KL{q(\xxi)}{p(\xxi)} = \frac{1}{2} \sum_{i=1}^{n_\xi} \left[
        \trace{\hat{\mtx S}_i^{(\xi)} - \log(\hat{\mtx S}_i^{(\xi)})}
        + \vc \mu_{i, :}^{(\xi)} \vc \mu_{i, :}^{(\xi) \intercal}
        - d_\xi
    \right],
\end{equation}
where $\hat{\mtx S}_i^{(\xi)} \in \reals^{d_\xi \times d_\xi}$ is a diagonal matrix with nonzero entries given by the $i$-th row of $\mtx S^{(\xi)}$ defined in Eq.\ (\ref{eqn:Theory:SGPLVM:qxi}), and
$\log(\hat{\mtx S}_i^{(\xi))})$ denotes the element-wise logarithm of $\hat{\mtx S}_i^{(\xi)}$, not its matrix logarithm~\cite{higham2008functions}.
Note that the collapsed bound of Eq.~(\ref{eqn:Theory:SGPLVM:CollapsedBound}) can also be rewritten in a way that exposes its factorization with respect to output dimensions as $\mc L = \sum_{j=1}^{d_y} \mc L_j - \KL{q(\xxi)}{p(\xxi)}$, with
\begin{equation}
    \begin{aligned}
        \mc L_j = &\frac{1}{2} \left(
            (n - m) \log \beta - n \log (2 \pi) - \log \determinant{\mtx A}
        \right)
        \\
        &- \frac{\beta}{2} \left(
            \trace{\vc y_{:, j} \vc y_{:, j}^\intercal}
            - \trace{\vc y_{:, j}^\intercal \mtx \Psi_1 \mtx K_\psi^{-1} \mtx \Psi_1^\intercal \vc y_{:, j}}
            + \psi_0 - \trace{\mtx C}
        \right).
    \end{aligned}
    \label{eqn:Theory:SGPLVM:CollapsedBoundDimension}
\end{equation}
To train the model, we maximize Eq.\ (\ref{eqn:Theory:SGPLVM:CollapsedBound}) using a gradient-based method over the variational parameters of Eq.~(\ref{eqn:Theory:SGPLVM:qxi}), the inducing inputs, and the model hyperparameters $\vc \theta = \{\vc \theta_k, \beta\}$.

\begin{remark}
    In order to emphasize the physical implications of our modeling choices---particularly, the inclusion of $\mtx X^{(s)}$ as explicit spatial inputs in addition to the usual latent variables $\mtx X^{(\xi)}$---we pause to clarify the meaning of the $d_y$ dimensions of the data in $\mtx Y$ to be modeled by way of an example.
    Consider the task of modeling $100$ color (RGB) images with a resolution of $640 \times 480$.
    In~\cite{damianou2016variational}, the authors discussed the use of a GP-LVM to model ``high-dimensional'' video; following their convention, one would represent this data set as a matrix $\mtx Y$ with $n=100$ rows (data) and $d_y = 3 \times 640 \times 480 \approx 9 \times 10^5$ columns (dimensions).
    In the current work, we conceptualize the same data set by defining $\mtx Y$ to have $n=100 \times 640 \times 480 = 3 \times 10^7$ rows (data) with $d_y = 3$ columns (dimensions).
    Thus, we model the data set not as ``few data, many dimensions'', but ``many data, few dimensions''.
    Since GPLVM models account for correlations between data, but not dimensions, the consequence of this distinction is that the SGPLVM immediately has a much richer model of the data set.
\end{remark}

\subsection{SGPLVM: efficient computation of the bound using structure-exploiting algebra}
\label{sec:Theory:SGPLVM:TrainingKronecker}
The drawback to the ``many data, few dimensions'' modeling choice is that it is computationally intractable if implemented na\"ively.
Next, we show how the structure in the model inputs can be exploited so that the time and memory requirements associated with training become linear in $n$.

First, noting that the training inputs are structured, we will likewise impose similar structure on the inducing points:
\begin{equation}
    \xu = \xuxi \times \xust,
    \label{eqn:Theory:SGPLVM:XuCartesianProduct}
\end{equation}
where $\xuxi \in \reals^{m_\xi \times d_\xi}$ and $\xust \in \reals^{m_s \times d_s}$.
This implies that 
\begin{align}
    \kuu &= \kuu^{(\xi)} \otimes \kuu^{(s)},
    \\
    \kfu &= \kfu^{(\xi)} \otimes \kfu^{(s)}.
\end{align}
While this assumption is not optimal, it produces a good working model in practice, is in a similar spirit to related works using structured inducing points approximations~\cite{wilson2015kernel, dai2017efficient}, and results in massive gains in computational efficiency that are necessary for tractability.

The statistics of Eq.\ (\ref{eqn:Theory:SGPLVM:PsiStatistics}) also simplify to
\begin{equation}
    \begin{aligned}
        \psi_0 &= \psi_0^{(\xi)} \trace{\kff^{(s)}},
        \\
        \psione &= \psione^{(\xi)} \otimes \kfu^{(s)},
        \\
        \psitwo &= \psitwo^{(\xi)} \otimes (\kfu^{(s)} \kuf^{(s)}),
    \end{aligned}
    \label{eqn:Theory:SGPLVM:PsiStatisticsStructured}
\end{equation}
where $\psi_0^{(\xi)}$, $\psione^{(\xi)}$, and $\psitwo^{(\xi)}$ may be evaluated in the usual way (i.e.\ using the formulae found in~\cite{damianou2015deep}).
The kernel over the stochastic input dimensions is subject to the same restrictions as usual if we wish for $\psione^{(\xi)}$ and $\psitwo^{(\xi)}$ to admit analytic forms.
Moreover, if the kernels $k_\xi$ and $k_s$ are stationary with variances $\sigma_{f,\xi}^2$ and $\sigma_{f, s}^2$, respectively, then
\begin{equation}
    \psi_0 = \sigma_{f, \xi}^2 \sigma_{f, s}^2 n.
\end{equation}

There are several immediate implications of these choices.
First, we notice that we are not required to compute any kernel expectations over the spatiotemporal dimensions of the inputs; this gives us extra flexibility in what kernels are available from an analytical/computational standpoint.
Second, by Eq.~(\ref{eqn:Theory:SGPLVM:XuCartesianProduct}), our model will have $m = m_\xi m_s$ inducing points in total, while the computational cost associated with evaluating the kernel expectations in Eq.~(\ref{eqn:Theory:SGPLVM:PsiStatisticsStructured}) remains unchanged relative to the traditional GP-LVM and still allows for parallelization with respect to the rows of $\xxi$ as discussed in~\cite{dai2014gaussian, gal2014distributed}.
To be more specific, we can write
\begin{align}
    (\mtx \Psi_1^{(\xi)})_{i, :} &= \expectation{\mtx K_{fu}^{(\xi)}}{q(\vc x_{i, :}^{(\xi)})}, ~ i = 1, \dots, n_\xi,
    \label{eqn:Theory:SGPLVM:PsiOneRows}
    \\
    \psitwo^{(\xi)} &= \sum_{i=1}^{n_\xi} 
    \expectation{(\kuf^{(\xi)})_{:, i} (\kfu^{(\xi)})_{i, :}}{q(\vc x_{i, :}^{(\xi)})}
    = \sum_{i=1}^{n_\xi} (\hat{\mtx \Psi}_2^{(\xi)})_i,
    \label{eqn:Theory:SGPLVM:PsiTwoSum}
\end{align}
where we see that each row in Eq.\ (\ref{eqn:Theory:SGPLVM:PsiOneRows}) and each summand in Eq.~(\ref{eqn:Theory:SGPLVM:PsiTwoSum}) depend on the marginal of a single training realization's latent variable.

\noindent
\begin{remark}
    If we choose $m_s = n_s$, then we can ``tie'' the corresponding inducing inputs to the training data's spatial points.
    This ansatz is motivated by the finding of Titsias~\cite{titsias2009variational} that, for sparse GPs with deterministic inputs, when $n=m$, then the optimal placement of the inducing points is such that $\mtx X = \mtx X_u$.
    Doing so also greatly reduces the number of parameters that must be optimized over and simplifies the computation of the variational bound since it implies that $\kff^{(s)} = \kfu^{(s)} = \kuu^{(s)}$.
    One may take a similar approach with the stochastic input space by tying the inducing points to the mode of $q(\xxi)$.
    However, this is not necessarily optimal in terms of maximizing the bound.
\end{remark}

Finally, we show how the remaining terms of the variational bound may be computed efficiently without having to ever explicitly evaluate any of the full $m \times m$ matrices in the bound.
Of course, we know from work on previous sparse GP models that no $n \times n$ matrices should ever have to be formed.
First, we note that $\mtx L$ and $\mtx C$ of Eqs.~(\ref{eqn:Theory:SGPLVM:L}) and~(\ref{eqn:Theory:SGPLVM:C}) have Kronecker decompositions:
\begin{align}
    \mtx L &= \mtx L^{(\xi)} \otimes \mtx L^{(s)},
    \\
    \mtx C &= \mtx C^{(\xi)} \otimes \mtx C^{(s)}.
\end{align}
Furthermore, we can use Eq.\ (\ref{eqn:Theory:SGPLVM:PsiTwoSum}) to obtain
\begin{equation}
    \mtx C^{(\xi)} = \sum_{i=1}^{n_\xi} (\hat{\mtx C}^{(\xi)})_i,
    \label{eqn:Theory:SGPLVM:CSum}
\end{equation}
The eigendecomposition of $\mtx C$ is
\begin{equation}
    \mtx C = \mtx Q_C \mtx \Lambda_C \mtx Q_C^\intercal,
\end{equation}
so $\mtx Q_C$ is orthogonal and $\mtx \Lambda_C$ is diagonal.
These both also admit Kronecker decompositions:
\begin{align}
    \mtx Q_C &= \mtx Q_C^{(\xi)} \otimes \mtx Q_C^{(s)},
    \label{eqn:Theory:SGPLVM:QcKronecker}
    \\
    \mtx \Lambda_C &= \mtx \Lambda_C^{(\xi)} \otimes \mtx \Lambda_C^{(s)}.
    \label{eqn:Theory:SGPLVM:LambdaCKronecker}
\end{align}
The matrices on the right hand sides of  Eqs.~(\ref{eqn:Theory:SGPLVM:QcKronecker}) and~(\ref{eqn:Theory:SGPLVM:LambdaCKronecker}) are found by computing the eigendecomposition of $\mtx C^{(\xi)}$ and $\mtx C^{(s)}$.
It follows that $\mtx A$ defined in Eq.\ (\ref{eqn:Theory:SGPLVM:A}) can be rewritten as
\begin{equation}
    \mtx A = \mtx Q_C \mtx D \mtx Q_C^\intercal.
    \label{eqn:Theory:SGPLVM:AEigen}
\end{equation}
where, for notational convenience, we have defined $\mtx D = \beta^{-1} \mtx I + \mtx \Lambda_C \in \reals^{m \times m}$.
Note that $\mtx D$ is the sum of two diagonals and thus contains $m$ nonzero entries.
Thus, it is at least as easy to store in memory as the data $\mtx Y$.
We also note that expressing $\mtx A$ as in Eq.~(\ref{eqn:Theory:SGPLVM:AEigen}) makes the computation of $\log \determinant{\mtx A}$ in Eq.~(\ref{eqn:Theory:SGPLVM:CollapsedBound}) straightforward:
\begin{equation}
    \log \determinant{\mtx A} = \sum_{i=1}^m \log d_{ii}.
\end{equation}
Next, we see that Eq.\ (\ref{eqn:Theory:SGPLVM:KPsi}) can be rewritten as
\begin{equation}
    \kpsi = \mtx L \mtx Q_C \mtx D \mtx Q_C^\intercal \mtx L^\intercal,
\end{equation}
and note that the matrix factors of
\begin{equation}
    \kpsiinv = \mtx L^{-\intercal} \mtx Q_C \mtx D^{-1} \mtx Q_C^\intercal \mtx L^{-1},
    \label{eqn:Theory:SGPLVM:KPsiInverse}
\end{equation}
may be efficiently computed and stored---$\mtx L$ and $\mtx Q_C$ are Kronecker products, meaning that their inverses are the Kronecker product of their submatrices' inverses; 
and the diagonal matrix $\mtx D$ can, of course, be inverted in an element-wise manner.

Finally, the second trace term in the second line of Eq.~(\ref{eqn:Theory:SGPLVM:CollapsedBound}) can be manipulated using Eq.~(\ref{eqn:Theory:SGPLVM:KPsiInverse}) to obtain
\begin{align}
    \trace{\mtx Y^\intercal \mtx \Psi_1 \mtx K_\psi^{-1} \mtx \Psi_1^\intercal \mtx Y} &= \trace{\mtx D^{-1} \underbrace{\mtx Q_C^\intercal \mtx L^{-1} \mtx \Psi_1^\intercal \mtx Y}_{\equiv \mtx B} \mtx Y^\intercal \mtx \Psi_1 \mtx L^{-\intercal} \mtx Q_C}
    \\
    &= \trace{\mtx D^{-1} \mtx B \mtx B^\intercal}
    \\
    &= \sum_{i=1}^n \sum_{j=1}^{d_y} d_{ii}^{-1} b_{ij}^2.
\end{align}
The matrix $\mtx B$ is most efficiently computed by first evaluating the product $\mtx Q_C^\intercal \mtx L^{-1} \mtx \Psi_1^\intercal$, then multiplying against $\mtx Y$ last.
Computing this term takes $\order{n d_y}$ time.

In summary, we have provided derivations for exploiting the structured inputs in the SGPLVM to obtain computationally-efficient ways of evaluating all of the terms in the collapsed lower bound of Eq.~(\ref{eqn:Theory:SGPLVM:CollapsedBound}).
Importantly, we see that computing the bound is linear in the size of the training data in both time and memory.
Training is carried out by optimizing Eq.~(\ref{eqn:Theory:SGPLVM:CollapsedBound}) using gradient-based methods.
Assuming that we have tied the spatial inducing inputs as described above, the parameters subject to optimization are the following:
\begin{itemize}
    \item Variational parameters: $\vc \mu^{(\xi)}$ and $\mtx S^{(\xi)}$, the variational parameters of $q(\xxi)$ of Eq.\ (\ref{eqn:Theory:SGPLVM:qxi}) ($2 n_\xi d_\xi$ parameters); and $\xuxi$, the latent variable inducing inputs ($m_\xi d_\xi$ parameters).
    \item Model hyperparameters: $\vc \theta_{k_\xi}$ and $\vc \theta_{k_s}$, the hyperparameters of the stochastic and spatial kernels (number of parameters depends on choice of kernels); and $\beta$, the precision of the Gaussian likelihood model ($1$ parameter).
\end{itemize}

\subsection{SPGLVM: predictions with the model}
\label{sec:Theory:SGPLVM:Predictions}
Having posed the variational lower bound for the model, we now turn our attention to using the trained model for prediction tasks.
We are interested with two tasks: propagating new latent variables through the posterior generative process to obtain samples in data space, and doing inference on new data to determine the corresponding posterior in latent space.
We address these in the following subsections.

\paragraph{Forward predictive density}
\label{sec:Theory:SGPLVM:Predictions:Forward}
Given some latent variables $\mtx X^* = \xxis \times \xsts \in \reals^{n^* \times d_x}$, predictions proceed by assuming that the inducing points are sufficient statistics of the training data.
The predictive density is therefore~\cite{damianou2016variational}
\begin{align}
    p(\mtx F^* | \mtx X^*) &= \prod_{j=1}^{d_{out}} \G{\vc f_j^*}{\vc \mu_j^*}{\mtx \Sigma^*},
    \label{eqn:Theory:SGPLVM:PredictiveDensity}
    \\
    \vc \mu^* &= \ksu \kpsi^{-1} \mtx \Psi_1^\intercal \mtx Y,
    \label{eqn:Theory:SGPLVM:x_to_y_mean}
    \\
    \mtx \Sigma^* &= \mtx K_{**} 
    - \ksu 
    \left( \kuuinv - \beta^{-1} \kpsi^{-1} \right) 
    \kus.
    \label{eqn:Theory:SGPLVM:x_to_y_covariance}
\end{align}
Computation of the predictive mean and variance are made efficient by exploiting the structure of the training and test inputs via Kronecker product properties.
First, we substitute Eq.~(\ref{eqn:Theory:SGPLVM:KPsiInverse}) into Eq.~(\ref{eqn:Theory:SGPLVM:x_to_y_mean}) to obtain
\begin{equation}
    \vc \mu^* = \ksu \mtx L^{-\intercal} \mtx Q_C \mtx D^{-1} \mtx Q_C^\intercal \mtx L^{-1} \mtx \Psi_1^\intercal \mtx Y.
    \label{eqn:Theory:SGPLVM:x_to_y_mean2}
\end{equation}
Second, we note that the test cross-covariance matrix $\ksu \in \reals^{n^* \times m}$, formed by computing the kernel function on combinations of the test and inducing inputs, has Kronecker structure:
\begin{equation}
    \ksu = \ksu^{(\xi)} \otimes \ksu^{(s)}.
\end{equation}
Putting these together, we see that Eq.~(\ref{eqn:Theory:SGPLVM:x_to_y_mean2}) may be efficiently computed for structured test inputs.
Importantly, we are able to use this model to predict at different spatiotemporal resolutions from that of our training data if desired.
This cannot be done without the parameterized kernel that captures spatiotemporal correlations in our model.

Next, we consider the variance $\diag(\mtx \Sigma^*)$.
Define $\mtx V^* \in \reals^{n^* \times d_y}$ such that $v_{ij}^*$ is the variance of $y_{ij}^*$.
We can see that $\vc v_{:, j}^* = \diag(\mtx \Sigma^*)$, $j = 1, \dots, d_y$.
By combining Eqs.~(\ref{eqn:Theory:SGPLVM:x_to_y_covariance}) and (\ref{eqn:Theory:SGPLVM:KPsiInverse}), we obtain
\begin{equation}
    \mtx \Sigma^* = \mtx K_{**} 
    - \ksu \mtx L^{-\intercal} \mtx Q_C \left(\mtx I - \beta^{-1} \mtx D^{-1} \right) \mtx Q_C^\intercal \mtx L^{-1}
    \kus,
    \label{eqn:Theory:SGPLVM:x_to_y_covariance_sub}
\end{equation}
and the diagonal can be computed as
\begin{equation}
    \diag (\mtx \Sigma^*) 
    = 
    \diag(\mtx K_{**}) 
    - \left( \ksu \mtx L^{-\intercal} \mtx Q_C \right) 
    \circ \left( \ksu \mtx L^{-\intercal} \mtx Q_C \right) 
    \diag \left(\mtx I - \beta^{-1} \mtx D^{-1} \right).
    \label{eqn:Theory:SGPLVM:PredictiveVariance}
\end{equation}
The full covariance of Eq.~(\ref{eqn:Theory:SGPLVM:x_to_y_covariance}) can be computed in $\order{m (n_s^{*})^2}$ time and $\order{(n_s^*)^2}$ memory; details are given in~\ref{apx:SGPLVM:PredictiveCovariance}.

If the test latent variable is described by a Gaussian posterior $q(\xxis)$, then marginalizing over it results in a non-analytic predictive density.
However, we can still compute its mean analytically:
\begin{equation}
    \bar{\vc \mu}^* = \expectation{\mtx F^*}{q(\xxis)} = \psione^* \kpsi^{-1} \mtx \Psi_1^\intercal \mtx Y,
    \label{eqn:Theory:SGPLVM:prediction:marginal_mean}
\end{equation}
where 
\begin{equation}
    \psione^* = \expectation{\ksu}{q(\xxis)} = \psione^{(\xi), *} \otimes \ksu^{(s)}.
\end{equation}
Next, we show how to approximate the variance of the marginal predictive density.
Since it is analytically intractable, we approximate it via sampling.
We do this by approximating the marginalized predictive density as a mixture of Gaussians.
First, we take $n_{MOG}$ samples from $q(\xxis)$:
\begin{equation}
    \hat{\mtx X}^{(\xi), *, i}, ~ i=1, \dots, n_{MOG}.
\end{equation}
Next, we compute the mean and variance, $\vc \mu^{*, i}$ and $\mtx V^{*, i}$ of the conditional Gaussian of Eq.~(\ref{eqn:Theory:SGPLVM:PredictiveDensity}) for each sample $\hat{\mtx X}^{(\xi), *, i}$.
Finally, the marginal density's variance is estimated as
\begin{equation}
    \bar{\mtx V}^* \approx \frac{1}{n_{MOG}} \sum_{i=1}^{n_{MOG}} 
    (\vc \mu^{*, i} - \bar{\vc \mu}^*) \circ (\vc \mu^{*, i} - \bar{\vc \mu}^*) + \mtx V^{*, i}.
    \label{eqn:Theory:SGPLVM:prediction:marginal_variance}
\end{equation}

\begin{remark}
The SGPLVM is a \textit{generative} model; one may use it to produce additional samples from the training data distribution.
This is done by first sampling a latent variable from the standard normal prior [Eq.~(\ref{eqn:Theory:SGPLVM:pxi})], then sampling a realization in data space from the forward density of Eq.~(\ref{eqn:Theory:SGPLVM:PredictiveDensity}).
However, the Gaussian process prior of Eq.~(\ref{eqn:Theory:GPR:Process}) at the core of the model implies that certain data may be easier to model than others, and limitations in the generative model's flexibility may invalidate the assumption that the latent variables associated with the true data distribution be standard normal-distributed in latent space.
We do not focus on the ability of the model to satisfy the standard normal prior in this work, though the question to what degree the SGPLVM model (or other models) can accurately model an unknown density in data space, given some finite set of examples, is of considerable interest.
\end{remark}

\paragraph{Inference of latent variables}
\label{sec:Theory:SGPLVM:Predictions:Backward}
Here, we explain how the SGPLVM can be used to infer the variational posterior over the latent variables associated with test observations.
Given some test observation $\mtx Y^* \in \reals^{n_s^* \times d_y}$ observed at spatial points $\xsts \in \reals^{n_s^* \times d_s}$, we would like to infer the posterior over its corresponding latent variable $\xxis \in \reals^{1 \times d_\xi}$:
\begin{equation}
    q(\vc x^{(\xi, *)}) = \prod_{j=1}^{d_\xi} \G{x_j^{(\xi), *}}{\mu_j^{(\xi), *}}{s_j^{(\xi), *}}.
\end{equation}
Note that the test observation might be observed at a different set of spatial points from the training data, and it may also have a different degree of noisiness.
The following methodology addresses both of these challenges.

First, we write the augmented variational lower bound for the training and test observations:
\begin{equation}
    \log p(\mtx Y^*, \mtx Y) \ge \mc L + \mc L^*,
    \label{eqn:Theory:SGPLVM:AugmentedBound}
\end{equation}
where the new term $\mc L^*$ follows the form of the uncollapsed bound of Eq.~(\ref{eqn:Theory:SGPLVM:UncollapsedBound}), leveraging the fact that it factorizes with respect to observations:
\begin{equation}
    \begin{aligned}
        \mc L^* = &-\frac{n^* d_y}{2} \left( \log(2 \pi) - \log \beta_* \right) 
        - \frac{\beta_*}{2} \trace{\mtx Y^* \mtx Y^{*, \intercal}}
        \\
        &- \frac{\beta_* d_y}{2} \trace{
            \kuu^{-1} \psitwo^* \kuu^{-1} (\Ubar \Ubar^\intercal + \mtx \Sigma_u)
        }
        + \beta_* \trace{\mtx Y^{*, \intercal} \psione^* \kuuinv \Ubar}
        \\
        &- \frac{\beta_* d_y}{2} \left( \psi_0^* - \trace{\kuu^{-1} \psitwo^*} \right)
        - \KL{q(\vc x^{(\xi, *)})}{p(\vc x^{(\xi, *)})},
    \end{aligned}
    \label{eqn:Theory:SGPLVM:BoundTestTerm}
\end{equation}
where $\Ubar$ and $\mtx \Sigma_u$ are given by Eq.~(\ref{eqn:Theory:SGPLVM:OptimalU}), and $\beta_*$ is the precision of the Gaussian observation likelihood corresponding to the test observation.
Furthermore, we can decompose $\mc L^*$ with respect to dimensions: $\mc L^* = \sum_{j=1}^{d_y} \mc F_j^* - \KL{q(\vc x^{(\xi, *)})}{p(\vc x^{(\xi, *)})}$, where
\begin{equation}
    \begin{aligned}
        \mc F_j^* = &-\frac{n^*}{2} \left( \log(2 \pi) - \log \beta_* \right) 
        - \frac{\beta_*}{2} \vc y_{:, j}^{*, \intercal} \vc y_{:, j}^*
        \\
        &- \frac{\beta_*}{2} \trace{
            \kuu^{-1} \psitwo^* \kuu^{-1} (\Ubar \Ubar^\intercal + \mtx \Sigma_u)
        }
        + \beta_* \vc y_{:, j}^{*, \intercal} \psione^* \kuuinv \ubar_{:, j}
        \\
        &- \frac{\beta_*}{2} \left( \psi_0^* - \trace{\kuu^{-1} \psitwo^*} \right)
    \end{aligned}
    \label{eqn:Theory:SGPLVM:BoundTestTermDimension}
\end{equation}
This approach may be seen as combining the ``collapsed'' bound of~\cite{titsias2010bayesian} to compute $\mc L$ and the ``uncollapsed'' bound for $\mc L^*$.
Note that in order for $q(\mtx U)$ to remain optimal, Eqs.~(\ref{eqn:Theory:SGPLVM:OptimalU:Mean}) and~(\ref{eqn:Theory:SGPLVM:OptimalU:Covariance}) would have to depend on the test realization.
However, this would eliminate the structure required for the model to be the computationally tractable.~\ref{apx:SGPLVM:TestInference} explains how to compute Eq.~(\ref{eqn:Theory:SGPLVM:BoundTestTerm}) efficiently.
To infer $q(\vc x^{(\xi, *)})$, one optimizes $\mc L + \mc L^*$ over its variational parameters.
In the interest of computational efficiency, we keep the variational parameters for the training data and kernel hyperparameters learned at training time constant.
Thus, $\mc L$ is constant while doing inference, and in practice our task simplifies to optimizing $\mc L^*$ alone over the $2 d_\xi$ variational parameters of $q(\vc x^{(\xi), *})$.
Note also that if we only have observations of $\mtx Y^*$ in certain dimensions, we can exploit the factorization of 
$\mc L^* = \mc F^* - \KL{q(\vc x^{(\xi, *)})}{p(\vc x^{(\xi, *)})}$
over output dimensions to optimize the bound on the marginal log-likelihood over the observed dimensions.
To do this, $\mc L^*$ is replaced by $\sum_{j \in \{\mc O\}} \mc F_j^* - \KL{q(\vc x^{(\xi, *)})}{p(\vc x^{(\xi, *)})}$, where $\{\mc O\}$ is the set of dimensions in which we have observations for $\mtx Y^*$.

\section{Application to inverse problems}
\label{sec:Theory:Inverse}
In this section, we describe how the proposed SGPLVM model can be used to solve inverse problems within a Bayesian formulation.

Consider some physical process $\tilde f(\tilde{\vc x}^{(\Xi)}, \vc x^{(s)})$ which maps some input parameters $\tilde{\vc x}^{(\Xi)}$ to an output at spatial location $\vc x^{(s)} \in \mc X^{(s)}$.
Physically, we assume in this work that $\tilde{\vc x}^{(\Xi)}$ is a scalar random field.
Furthermore, we assume that we have access to some approximation of the physical process through a computer code (e.g.\ finite element solver) that solves a set of partial differential equations, which we will denote here as $f(\vc x^{(\Xi)}, \vc x^{(s)})$, where $\vc x^{(\Xi)}$ is a finite-dimensional projection of $\tilde{\vc x}^{(\Xi)}$ from discretization of the problem.
We can compute the forward model $f$ at any provided input $\vc x^{(\Xi)} \in \mc X^{(\Xi)}$, but doing so is assumed to be computationally expensive.

Lastly, we are provided with a set of noisy measurements of $\tilde f$ at a set of spatial locations $\{ \vc x_i^{(s)}\}_{i=1}^{n_s^*}$, represented as a vector $\tilde{\vc y}$.
Given a stochastic prior model $p(\vc x^{(\Xi)})$, our objective is to determine the posterior $p(\vc x^{(\Xi)} | \tilde{\vc y})$.
Here, we will consider the case of a data-driven prior characterized by a collection of realizations of the random field.
Such a scenario might occur when one has access to a series of calibration measurements on a known object or samples from a trusted physical model of the input, and is critical to building strong priors in a variety of domain applications such as materials modeling and medical analysis~\cite{ganapathysubramanian2008non, kaipio2006statistical}.

The Bayesian inverse problem thus posed has two notable challenges.
First, the stochastic input $\vc x^{(\Xi)}$ is high-dimensional, meaning that straightforward exploration of the stochastic space is unlikely to be effective in practice.
Thus, while some sort of dimensionality reduction of the empirical input data set is needed, we require that it be able to accurately capture the epistemic uncertainty due to the finite nature of the data set.
Additionally, the dimensionality reduction model must be generative so that one may go from latent space back to the input data space.
Second, evaluating the forward model is computationally expensive, limiting how many times we may invoke it.
A surrogate model is needed to alleviate the computational burden; however, we require a methodology that rigorously propagates the uncertainty in the learned latent variables from our dimensionality reduction.
The SGPLVM provides a means of solving these challenges, as we now explain.
We will discuss two possible approaches: a ``two-model'' approach in which two SGPLVM ``submodels'' are used for the problem inputs and outputs, and a ``joint model'' approach in which a representation of inputs and outputs is learned simultaneously.

In order to train the SGPLVM models, let us assume that we have $n_{\Xi, in}$ realizations from the stochastic prior model, each observed on a set of $n_{s, in}$ spatial locations.\footnote{In our discussion of the inverse problem and subsequent experiments, we will append either ``in'' or ``out'' to the variables to indicate that they belong to either the input or output data and/or SGPLVM submodels as applicable if they are not equal.}
Of these, $n_{\Xi, out} \le n_{\Xi, in}$ have been solved using our forward model, returning an output at $n_{s, out}$ spatial locations.
Anticipating their use within the SGPLVM, the training inputs are assembled into a matrix $\mtx Y^{(in)} \in \reals^{n_{in} \times 1}$, where $n_{in} = n_{\Xi, in} n_{s, in}$.

\subsection{Two-model approach}
The two-model approach is comprised of two SGPLVMs: an ``input submodel'' and an ``output submodel''.
The input submodel is trained on the realizations of the stochastic inputs alone to learn a low-dimensional latent representation.
Given this latent representation, the output submodel subsequently acts as a surrogate to the forward model (FEM solver), learning a mapping from the uncertain latent variables to the corresponding outputs.

In terms of the SGPLVM formulation, the input submodel is provided with $\mtx Y^{(in)}$ and the spatial locations $\mtx X^{(s, in)}$ as training data.
Training yields a variational posterior over the training examples' latent variables, $q(\mtx X^{(\xi, in)})$.
Note that the number of stochastic realizations for the input submodel is just $n_{\xi, in} = n_{\Xi, in}$.

\begin{remark}
    A more traditional approach to dimensionality reduction might be to compute a Karhunen-Lo\`eve expansion on the input realizations.  
    The latent variables here fulfill the same role as the KL coefficients with the exception that they are distributions due to the data-driven nature of the dimensionality reduction, and the mapping from the latent variables to data space is given as a GP and is therefore probabilistic and nonlinear.
\end{remark}

Next, we train the output submodel.
To do this, we fix the latent variable posterior $q(\mtx x^{(\xi, out)})$ to be equal to $q(\mtx x^{(\xi, in)})$ after removing any columns of $\mtx X^{(\xi, in)}$ corresponding to inputs that have not been solved on.
Training the output submodel optimizes over only the kernel hyperparameters and inducing points; the latent variables must remain fixed to ensure that the input and output submodels possess a consistent latent representation of input-output pairs.
The training procedure is summarized in Algorithm~\ref{alg:Inverse:2Model:Train}.
Figure~\ref{fig:Theory:Inverse:Schematic:2Model} shows a schematic of this modeling approach.

\begin{algorithm}
    \textbf{Require:} Training input and output data $\mtx X^{(s, in)}$, $\mtx Y^{(in)}$, $\mtx X^{(s, out)}$, and $\mtx Y^{(out)}$.
    
    \textbf{Ensure:} Trained input and output SGPLVM submodels.
    
    \begin{algorithmic}[1]
        \State Provide $\mtx Y^{(in)}$ and $\mtx X^{(s, in)}$ as training data to the input SGPLVM submodel.
        \State Train the input submodel by optimizing the lower bound of Eq.~(\ref{eqn:Theory:SGPLVM:CollapsedBound}) over all variational parameters and model hyperparameters.
        \State Copy the variational posterior means and variances $\vc \mu^{(\xi, in)}$ and $\mtx S^{(\xi, in)}$ to the output submodel to define the latent variable posterior $q(\mtx X^{(\xi, out)})$, eliminating any rows corresponding to input cases that were not solved using the simulator.
        \State Provide $\mtx Y^{(out)}$ and $\mtx X^{(s, out)}$ as training data to the output SGPLVM submodel as well as $q(\mtx X^{(\xi, out)})$.
        \State Train the output submodel by optimizing the lower bound of Eq.~(\ref{eqn:Theory:SGPLVM:CollapsedBound}) over $\mtx X_u$ and the model hyperparameters.
    \end{algorithmic}
    \caption{Training procedure for the two-model approach.}
    \label{alg:Inverse:2Model:Train}
\end{algorithm}

Given the trained submodels, we may now make predictions.
Note that, due to the symmetry of the setup, the procedure for traditional ``forward'' predictions of the simulator output at some new test input are almost the same as the reverse prediction in an inverse problem.
However, the generative model's capability of modeling noisy observations as well as propagating the uncertainty of projecting into latent space as well as uplifting back to data space will ensure that predictions possess well-calibrated uncertainty estimates reflecting the model's epistemic uncertainty as well as the ill-posed nature of the inverse problem.

In either case, predictions are carried out in two steps.
First, we consider the case of predicting the simulator output, given a new test input from the stochastic space, $\vc x^{(\Xi)}$.
The first step is to infer the posterior $q(\vc x^{(\xi, in), *})$ for the test case; 
this is done using the input submodel.
We first reshape $\vc x^{(\Xi), *}$ as necessary to obtain $\mtx Y^{(in), *}$.
We then optimize $\mc L^{in} + \mc L^{in, *}$ of Eq.~(\ref{eqn:Theory:SGPLVM:BoundTestTerm}) over the variational parameters of $q(\vc x^{(\xi, in), *})$.
Next, we use the output submodel to predict the output by computing the forward predictive density from the test input $q(\vc x^{(\xi, out), *}) = q(\vc x^{(\xi, in), *})$ as discussed in Section~\ref{sec:Theory:SGPLVM:Predictions:Forward} using the output submodel.
The procedure is summarized in Algorithm~\ref{alg:Inverse:2Model:PredictForward}.

\begin{algorithm}
    \textbf{Require:} Trained SGPLVM input and output submodels, input realization $\vc x^{(\Xi), *}$ observed at spatial points $\mtx X^{(s, in), *}$, and spatial points $\mtx X^{(s, out), *}$ at which the output is to be predicted.
    
    \textbf{Ensure:} Mean and variance of the posterior over the output.
    
    \begin{algorithmic}[1]
        \State Reshape $\vc x^{(\Xi), *}$ to obtain $\mtx Y^{(in), *} \in \reals^{n_{s, in}^* \times 1}$. 
        \State Infer $q(\vc x^{(\xi, in), *})$ by optimizing $\mc L^{in} + \mc L^{in, *}$ over $\vc \mu^{(\xi, in), *}$ and $\vc s^{(\xi, in), *}$ using the input submodel.
        \State Set $q(\vc x^{(\xi, out), *}) = q(\vc x^{(\xi, in), *})$
        \State Compute $\bar{\vc \mu}^*$ using Eq.~(\ref{eqn:Theory:SGPLVM:prediction:marginal_mean}) and $\bar{\mtx V}^*$ using Eq.~(\ref{eqn:Theory:SGPLVM:prediction:marginal_variance}) with the output submodel.
    \end{algorithmic}
    \caption{Forward predictions for the two-model approach.}
    \label{alg:Inverse:2Model:PredictForward}
\end{algorithm}

In the inverse case, we are provided instead with noisy observations $\tilde{\vc y}$ observed on $\mtx X^{(s, out), *}$ as mentioned above.
However, the prediction procedure is rather similar.
First, we reshape $\tilde{\vc y}$ as necessary to obtain $\mtx Y^{(out), *}$.
We then infer $q(\vc x^{(\xi, out), *})$ again using the partially collapsed bound with the output submodel.
In this case, we additionally optimize over $\beta_*$ instead of fixing it to $\beta$ in order to correctly infer the noisiness of the provided test data.
Lastly, the posterior over the stochastic input $\vc x^{(\Xi), *}$ is computed by propagating $q(\vc x^{(\xi, in), *}) = q(\vc x^{(\xi, out), *})$ through the forward predictive density of Eq.~(\ref{eqn:Theory:SGPLVM:PredictiveDensity}) using the input submodel.
The procedure is summarized in Algorithm~\ref{alg:Inverse:2Model:PredictInverse}.

\begin{algorithm}
    \textbf{Require:} Trained SGPLVM input and output submodels, output realization $\tilde{\vc y}$ observed at spatial points $\mtx X^{(s, out), *}$, and spatial points $\mtx X^{(s, in), *}$ at which the input is to be inferred.
    
    \textbf{Ensure:} Mean and variance of the posterior over the input.
    
    \begin{algorithmic}[1]
        \State Reshape $\tilde{\vc y}$ to obtain $\mtx Y^{(out), *} \in \reals^{n_{s, out}^* \times 1}$. 
        \State Infer $q(\vc x^{(\xi, out), *})$ by optimizing $\mc L^{out} + \mc L^{out, *}$ over $\vc \mu^{(\xi, out), *}$, $\vc s^{(\xi, out), *}$, and $\beta_*$ using the output submodel.
        \State Set $q(\vc x^{(\xi, in), *}) = q(\vc x^{(\xi, out), *})$.
        \State Compute $\bar{\vc \mu}^*$ using Eq.~(\ref{eqn:Theory:SGPLVM:prediction:marginal_mean}) and $\bar{\mtx V}^*$ using Eq.\ (\ref{eqn:Theory:SGPLVM:prediction:marginal_variance}) with the input submodel.
    \end{algorithmic}
    \caption{Inverse predictions for the two-model approach.}
    \label{alg:Inverse:2Model:PredictInverse}
\end{algorithm}

\subsection{Jointly-trained model approach}
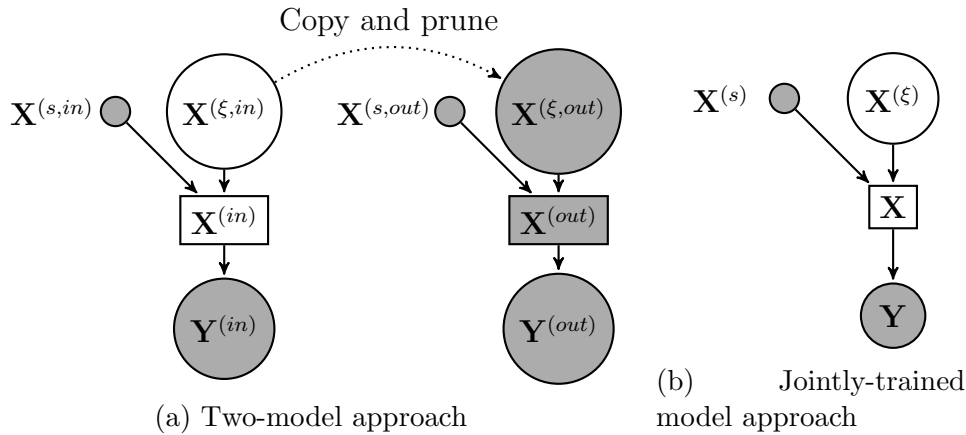
\begin{figure}
    \centering
    \begin{subfigure}[b]{.65\linewidth}
        \centering
\begin{tikzpicture}[->,
    >=stealth',
    shorten >=1pt,
    auto,
    node distance=3.5em, 
    thick,
    observed/.style={
        circle,
        minimum size=0.05cm,
        draw,
        fill=gray!65
    },
    unobserved/.style={
        circle,
        minimum size=0.05cm,
        draw
    },
    deterministic_unobserved/.style={
        rectangle,
        draw
    },
    deterministic_observed/.style={
        rectangle,
        draw,
        fill=gray!65
    }
    ] 
    \node[unobserved]    (x_xi_in) {$\mtx X^{(\xi, in)}$};
    \node[observed]      (x_st_in) [left of=x_xi_in] {};
    \node                (x_st_in_label) [left of=x_st_in, xshift=0.6cm] {$\mtx X^{(s, in)}$};
    \node[deterministic_unobserved] (x_in) [below of=x_xi_in]  {$\mtx X^{(in)}$};
    \node[observed]      (y_in) [below of=x_in] {$\mtx Y^{(in)}$};
    
    \node[observed]      (x_xi_out) [right of=x_xi_in, xshift=3cm] {$\mtx X^{(\xi, out)}$};
    \node[observed]      (x_st_out) [left of=x_xi_out] {};
    \node                (x_st_out_label) [left of=x_st_out, xshift=0.57cm] {$\mtx X^{(s, out)}$};
    \node[deterministic_observed] (x_out) [below of=x_xi_out]  {$\mtx X^{(out)}$};
    \node[observed]      (y_out) [below of=x_out] {$\mtx Y^{(out)}$};

    \path
        (x_xi_in) edge (x_in)
        (x_st_in) edge (x_in)
        (x_in) edge (y_in)
        
        (x_xi_out) edge (x_out)
        (x_st_out) edge (x_out)
        (x_out) edge (y_out)
        ;
        
    \path(x_xi_in) edge[dotted, bend left] node {Copy and prune} (x_xi_out);
\end{tikzpicture}

        \caption{Two-model approach}
        \label{fig:Theory:Inverse:Schematic:2Model}
    \end{subfigure}
    \begin{subfigure}[b]{.3\linewidth}
        \centering
\begin{tikzpicture}[->,
    >=stealth',
    shorten >=1pt,
    auto,
    node distance=3.5em, 
    thick,
    observed/.style={
        circle,
        minimum size=0.05cm,
        draw,
        fill=gray!65
    },
    unobserved/.style={
        circle,
        minimum size=0.05cm,
        draw
    },
    deterministic/.style={
        rectangle,
        draw
    }
    ] 
    \node[unobserved]    (x_xi_in) {$\mtx X^{(\xi)}$};
    \node[observed]      (x_st_in) [left of=x_xi_in] {};
    \node                (x_st_in_label) [left of=x_st_in, xshift=0.6cm] {$\mtx X^{(s)}$};
    \node[deterministic] (x_in) [below of=x_xi_in]  {$\mtx X$};
    \node[observed]      (y_in) [below of=x_in] {$\mtx Y$};

    \path
        (x_xi_in) edge (x_in)
        (x_st_in) edge (x_in)
        (x_in) edge (y_in)
        ;
\end{tikzpicture}

        \caption{Jointly-trained model approach}
        \label{fig:Theory:Inverse:Schematic:JointModel}
    \end{subfigure}
    \caption{
        Illustration of the two modeling approaches.
        In \subref{fig:Theory:Inverse:Schematic:2Model}, we first train an ``input submodel'' to learn a latent representation for the input realizations, then provide the learned posterior as uncertain, but fixed inputs to the ``output submodel'', which learns a regression to the corresponding solutions.
        In \subref{fig:Theory:Inverse:Schematic:JointModel}, we treat the input and output as different dimensions and jointly learn a latent representation using a  single SGPLVM for both simultaneously.
        Inducing variables, latent outputs, and model hyperparameters included in the model as per Fig.~\ref{fig:Theory:SGPLVM:PGM} are not shown in the interest of clarity.
    }
    \label{fig:Theory:Inverse:Schematic}
\end{figure}

In the jointly-trained model approach, we will train a single SGPLVM model that simultaneously learns a shared latent representation of the inputs and outputs.
Figure~\ref{fig:Theory:Inverse:Schematic:JointModel} shows a schematic of this modeling approach.
To do this, we first concatenate the input and output data matrices $\mtx Y^{(in)}$ and $\mtx Y^{(out)}$ to form $\mtx Y = (\mtx Y^{(in)}, \mtx Y^{(out)}) \in \reals^{n \times 2}$.
In order to ensure that this is possible, we must satisfy several restrictions on the training data to be modeled.
First, we require that $\mtx X^{(s)} = \mtx X^{(s, in)} = \mtx X^{(s, out)}$, and thus that $n_{s, in} = n_{s, out}$.
Second, we require that $n_{\xi, in} = n_{\xi, out}$, i.e., that all input realizations to be trained on have a corresponding solution computed.
These two criteria imply that $n_{in} = n_{out} = n$.

Training is achieved by a straightforward optimization of the variational bound for the single SGPLVM model as is summarized in Algorithm~\ref{alg:Inverse:JointModel:Train}.
The parameters subject to optimization are itemized in Section~\ref{sec:Theory:SGPLVM:TrainingKronecker}.

\begin{algorithm}
    \textbf{Require:} Training input and output data $\mtx X^{(s)}$, $\mtx Y^{(in)}$, and $\mtx Y^{(out)}$.
    
    \textbf{Ensure:} Trained SGPLVM model.
    
    \begin{algorithmic}[1]
        \State Provide $\mtx Y = (\mtx Y^{(in)}, \mtx Y^{(out)})$ and $\mtx X^{(s)}$ as training data to the SGPLVM model.
        \State Train the model by optimizing the lower bound of Eq.~(\ref{eqn:Theory:SGPLVM:CollapsedBound}) over all variational parameters and model hyperparameters.
    \end{algorithmic}
    \caption{Training procedure for the jointly-trained approach.}
    \label{alg:Inverse:JointModel:Train}
\end{algorithm}

As in the two-model approach, both forward and inverse predictions are handled in a very similar way and again involve a two-step procedure.
We describe the forward prediction procedure here.
We first reshape $\vc x^{(\Xi), *}$ as necessary to obtain $\mtx Y^{(in), *}$.
This is the first column of the test observation $\mtx Y^* \in \reals^{n_s^* \times 2}$.
We then infer the variational posterior $q(\vc x^{(\xi), *})$ by exploiting the factorization of $\mc L^*$ over dimensions and optimizing the quantity 
$\mc F^* - \KL{q(\vc x^{(\xi), *})}{p(\vc x^{(\xi), *})}$ over the variational parameters of $q(\vc x^{(\xi), *})$.
Next, we compute the forward predictive density over the output dimension based on $q(\vc x^{(\xi), *})$ as discussed in 
Section~\ref{sec:Theory:SGPLVM:Predictions:Forward}.
The procedure is summarized in Algorithm~\ref{alg:Inverse:JointModel:PredictForward}.

\begin{algorithm}
    \textbf{Require:} Trained SGPLVM, input realization $\vc x^{(\Xi), *}$ observed at spatial points $\mtx X^{(s, in), *}$, and spatial points $\mtx X^{(s, out), *}$ at which the output is to be predicted.
    
    \textbf{Ensure:} Mean and variance of the posterior over the output.
    
    \begin{algorithmic}[1]
        \State Reshape $\vc x^{(\Xi), *}$ to obtain $\mtx Y^{(in), *} \in \reals^{n_{s, in}^* \times 1}$. 
        \State Infer $q(\vc x^{(\xi), *})$ by optimizing $\mc L + \mc F_1^* - \KL{q(\vc x^{(\xi), *})}{p(\vc x^{(\xi), *})}$ over $\vc \mu^{(\xi), *}$ and $\vc s^{(\xi), *}$.
        \State Compute $\bar{\vc \mu}_{:, 2}^*$ using Eq.~(\ref{eqn:Theory:SGPLVM:prediction:marginal_mean}) and $\bar{\vc v}_{:, 2}^*$ using Eq.~(\ref{eqn:Theory:SGPLVM:prediction:marginal_variance}).
    \end{algorithmic}
    \caption{Forward predictions for the jointly-trained approach.}
    \label{alg:Inverse:JointModel:PredictForward}
\end{algorithm}

As with the two-model approach, the inverse prediction is accomplished by simply switching the roles of the two data dimensions.
Again, we optimize over $\beta_*$ when solving the inverse problem to account for the fact that we do not expect the noise in the test data to match that of the forward model solution data used for training.
The approach is summarized in Algorithm~\ref{alg:Inverse:JointModel:PredictInverse}.

\begin{algorithm}
    \textbf{Require:} Trained SGPLVM, output realization $\tilde{\vc y}$ observed at spatial points $\mtx X^{(s, out), *}$, and spatial points $\mtx X^{(s, in), *}$ at which the input is to be inferred.
    
    \textbf{Ensure:} Mean and variance of the posterior over the input.
    
    \begin{algorithmic}[1]
        \State Reshape $\tilde{\vc y}$ to obtain $\mtx Y^{(out), *} \in \reals^{n_{s, out}^* \times 1}$. 
        \State Infer $q(\vc x^{(\xi), *})$ by optimizing $\mc L + \mc F_2^* - \KL{q(\vc x^{(\xi), *})}{p(\vc x^{(\xi), *})}$ over $\vc \mu^{(\xi), *}$, $\vc s^{(\xi), *}$, and $\beta_*$.
        \State Compute $\bar{\vc \mu}_{:, 1}^*$ using Eq.\ (\ref{eqn:Theory:SGPLVM:prediction:marginal_mean}) and $\bar{\vc v}_{:, 1}^*$ using Eq.~(\ref{eqn:Theory:SGPLVM:prediction:marginal_variance}).
    \end{algorithmic}
    \caption{Inverse predictions for the jointly-trained approach.}
    \label{alg:Inverse:JointModel:PredictInverse}
\end{algorithm}

There are advantages and disadvantages to the two different approaches described above.
The main advantage of the two-model approach over the jointly-trained model is that we are free to select the spatial points independently for each submodel, and we do not need to have solutions for all of our inputs in order to leverage them in the input submodel.
Additionally, we do not require that the solution has the same spatial structure; such cases are not uncommon with adaptive solver or mixed-element methods.
Furthermore, the two-model approach could allow one to leverage the learned latent space from the input submodel to strategically pick which inputs to solve in order to obtain an efficient experimental design that will help in creating a high-quality output submodel.
On the other hand, the jointly-trained approach will create a latent representation with the goal of being able to generate both inputs and outputs in a balanced way, whereas the two-model approach takes a greedy approach to learning the latent space.  
This should result in greater data efficiency than the two-model approach, all other things being equal.
This is particularly the case for inverse predictions, where optimizing the latent space to generate the output data can be expected to improve the SGPLVM's capabilities when inferring latent variables from noisy test outputs.

\section{Examples}
\label{sec:Examples}

We will now apply the methodology developed in Section \ref{sec:Theory} to a stochastic elliptic PDE with an uncertain conductivity field.
Code and data necessary for replicating the following experiments will be made available through a GitHub repository.\footnote{\url{https://github.com/cics-nd/sgplvm-inverse}}
We will consider the following elliptic SPDE:
\begin{equation}
    \begin{aligned}
        -\nabla \cdot \left( a_k(\vc x^{(s)}, \vc \omega) \nabla u(\vc x^{(s)}, \vc \omega) \right) &= 0 ~\textrm{in} ~ \mc X_s \times \mc X_\omega,
        \nonumber
        \\
        u(\vc x, \vc \omega) &= 1 - x_{s, 1} ~ \textrm{on} ~ \Gamma_0 \times \mc X_\omega,
        \nonumber
        \\
        \nabla_{\vc x^{(s)}} u(\vc x, \vc \omega) \cdot \hat{\vc n} &= 0 ~ \textrm{on} ~ \Gamma_n \times \mc X_\omega,
        \nonumber
    \end{aligned}
    \label{eqn:Elliptic}
\end{equation}
with 
$\vc x^{(s)} \in \mc X_s = [0, 1]^2$, 
$\Gamma_0 = \{\vc x^{(s)}: x_1^{(s)}=0 ~ \textrm{or} ~ x_1^{(s)}=1 \}$, and 
$\Gamma_n = \{\vc x^{(s)}: x_2^{(s)}=0 ~ \textrm{or} ~ x_2^{(s)}=1 \}$.
The conductivity $a_k$ is a random field; we will not use an explicit parameterization of it (e.g.\ KL expansion) but instead learn its structure through a set of example realizations at $n_s = \tilde{n}_s^2$ grid points ($\tilde{n}_s = 65$, so $n_s=4225$).

\begin{figure}
    \centering
    \includegraphics[height=\textwidth, angle=-90]{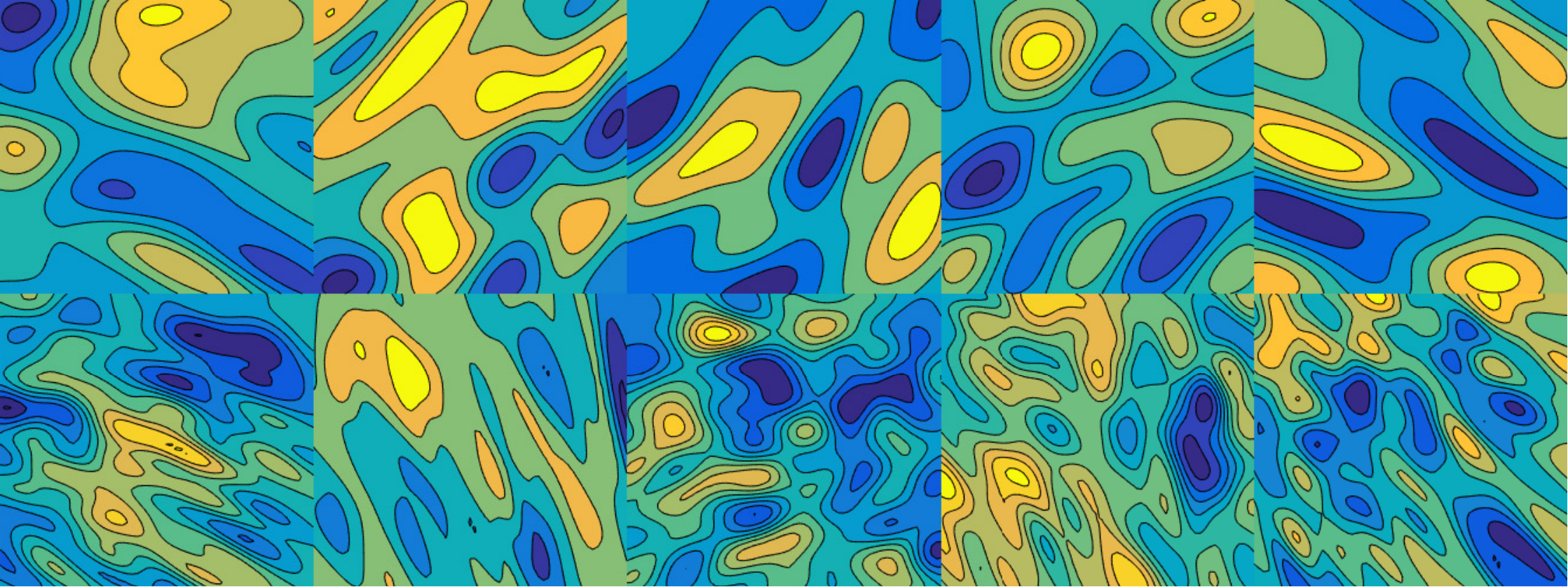}
    \caption{Sample inputs with $d_{KL} = (16, 32)$ (top) and $(16, 128)$ (bottom).}
    \label{fig:SGPLVM:EllipticForward:SampleInputs}
\end{figure}

The stochastic input model is defined implicitly through a two-layer (warped) GP as follows:
\begin{align}
    {\vc x^{(s)}}'(\vc x^{(s)}) &\sim \GP{\vc \mu}{k_1(\cdot, \cdot; \theta_1)},
    \label{eqn:Examples:Elliptic:GP1}
    \\
    \log a_k({\vc x^{(s)}}') &\sim \GP{0}{k_2(\cdot, \cdot; \theta_2)},
    \label{eqn:Examples:Elliptic:GP2}
\end{align}
where the mean function and kernels are chosen as follows:
\begin{align}
    \vc \mu(\vc x^{(s)}) &= \vc x^{(s)},
    \nonumber
    \\
    k_1(\vc x_{i, :}, \vc x_{j, :}) &= \sigma_{k, 1}^2 \exp \left[ -\sum_{k=1}^{d_s} \left( \frac{x_{ik} - x_{jk}}{l_1}\right)^2 \right],
    \nonumber
    \\
    k_2(\vc x_{i, :}', \vc x_{j, :}') &= \sigma_{k, 2}^2 \exp \left( -\norm{\frac{x_{i, :}' - x_{j, :}'}{l_2}} \right),
    \nonumber
\end{align}
with $\sigma_{k, 1}^2 = 0.25$, $\sigma_{k, 2}^2 = 1$, $l_1 = 2$, and $l_2 = 0.1$.
The Gaussian processes of Eqs.~(\ref{eqn:Examples:Elliptic:GP1}) and~(\ref{eqn:Examples:Elliptic:GP2}) are approximated with Karhunen-Lo\`eve expansions with $d_{KL, 1}$ and $d_{KL, 2}$ terms:
\begin{align}
    {\vc x^{(s)}}'(\vc x^{(s)}) &\approx \vc \mu(\vc x^{(s)}) + 
    \sum_{i=1}^{d_{KL, 1}} \omega_i^{(1)} \lambda_i^{(1)} \vc \phi^{(1, i)}(\vc x^{(s)}), ~ 
    &\vc \omega^{(1)} \sim \mc N(0, \mtx I_{d_{KL, 1} \times d_{KL, 1}}),
    \label{eqn:Elliptic:KL1}
    \\
    \log a_k({\vc x^{(s)}}') &\approx 
    \sum_{i=1}^{d_{KL, 2}} \omega_i^{(2)} \lambda_i^{(2)} \vc \phi^{(2, i)}({\vc x^{(s)}}'), ~ 
    &\vc \omega^{(2)} \sim \mc N(0, \mtx I_{d_{KL, 2} \times d_{KL, 2}}).
    \label{eqn:Elliptic:KL2}
\end{align}
In this work, we use $d_{KL, 1} = 16$ and consider both $d_{KL, 2} = 32$ and $128$.
This implicitly defines a stochastic probability model; Algorithm~\ref{alg:Elliptic:Input} explains how one can produce samples from it.
Importantly, the surrogate that we will construct does not have access to the random coefficients used in the KL expansions, but rather learns its representation from the samples of the input field themselves.
Example realizations of the generative process are shown in Fig.~\ref{fig:SGPLVM:EllipticForward:SampleInputs}.
\begin{algorithm}
    \textbf{Require:} Variances $\sigma_{k, 1}, \sigma_{k, 2}$ and length scales $l_1, l_2$, the number of KL terms for each layer $d_{KL, 1}, d_{KL, 2}$, and spatial points $\mtx X^{(s)}$ on which discrete samples will be generated.
    
    \textbf{Ensure:} Samples from the implicit stochastic prior model.
    
    \begin{algorithmic}[1]
        \State Compute the KL expansion of the GP of Eq.~(\ref{eqn:Examples:Elliptic:GP1}) on $\mtx X^{(s)}$ and truncate it to $d_{KL, 1}$ terms $\{\lambda_i^{(1)}, \vc \phi^{(1, i)}\}_{i=1}^{d_{KL, 1}}$.
        \For{$t = 1, \dots, n_\xi$}
            \State Sample $\vc \omega^{(1)}$ and compute ${\vc x^{(s)}}'(\vc x^{(s)})$ using Eq.~(\ref{eqn:Elliptic:KL1}).  Assemble it into ${\mtx X^{(s)}}' \in \reals^{n_s \times d_s}$
            \State Compute the KL expansion of the GP of Eq.~(\ref{eqn:Examples:Elliptic:GP2}) on ${\mtx X^{(s)}}'$ and truncate to $d_{KL, 2}$ terms $\{\lambda_i^{(2)}, \vc \phi^{(2, i)}\}_{i=1}^{d_{KL, 2}}$.
            \State Sample $\vc \omega^{(2)}$ and compute $\log a_k({\vc x^{(s)}}')$ using Eq.~(\ref{eqn:Elliptic:KL2}).
            \State The $t$-th sample of $a_k(\vc x^{(s)})$ is given as $\exp(\log a_k)$.
        \EndFor
    \end{algorithmic}
    \caption{Sampling from the stochastic prior model.}
    \label{alg:Elliptic:Input}
\end{algorithm}

This implicit stochastic prior model process is broader than the typical GP prior model over $\log a_k$ commonly found in the literature (see for example,~\cite{marzouk2009dimensionality, bilionis2013solution}) in that the warping GP of Eq.~(\ref{eqn:Examples:Elliptic:GP1}) allows us to consider not only different length scales from realization to realization, but varying length scales \textit{within} each realization within our stochastic prior model.
The traditional assumption that the length scale is known exactly \textit{a priori} has been criticized as being overly restrictive; see, for example, the work of Tripathy and Bilionis~\cite{tripathy2018deep}, who advocate for data-driven approaches to modeling the high-dimensional input space, as we do here.

\begin{figure}[hbt]
    \centering
    \includegraphics[width=\textwidth]{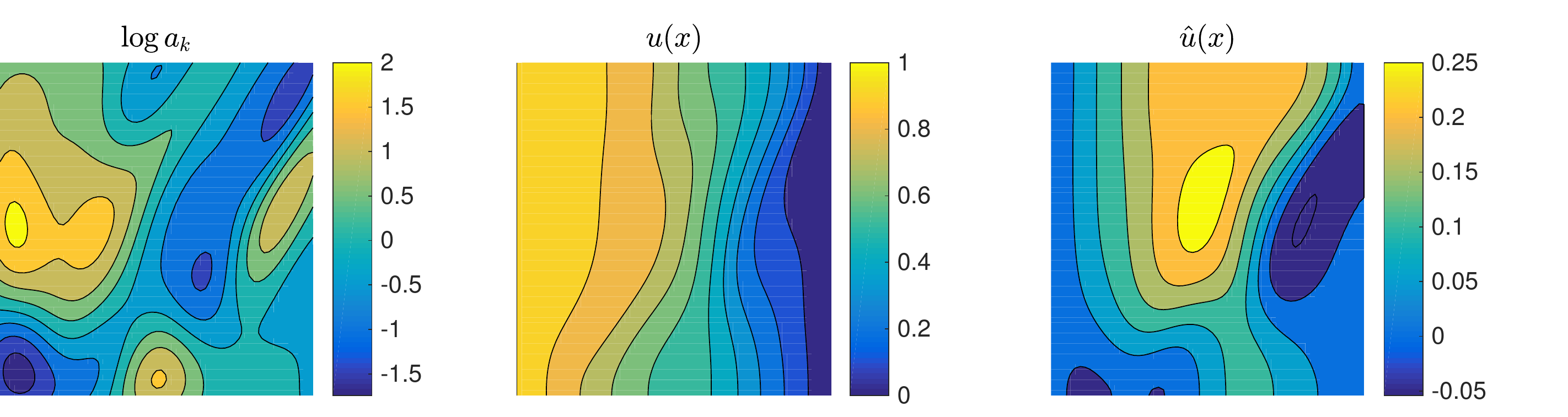}
    \caption{Sample input (left), solution $u(\vc x^{(s)})$, and solution with the mean response removed $\hat{u} = u - (1 - x_{s, 1})$.}
    \label{fig:Examples:SGPLVM:EllipticForward:SampleSolve}
\end{figure}

The PDE is solved, given an input realization, using FEM over a uniform triangular mesh with $2((\tilde{n}-1)^2)$ elements and $n_s = \tilde{n}_s^2$ nodes.
We will concern ourselves with the discrete solution at the nodes ($\mtx X^{(s)} \in \reals^{n_s \times d_s}$).
An example solution is shown in Fig.~\ref{fig:Examples:SGPLVM:EllipticForward:SampleSolve}.

\subsection{Forward surrogate model}
\label{sec:Examples:Elliptic:Forward}
We explore using both the two-model and jointly-trained model approaches for the this problem.
Training data is produced by sampling the stochastic input model via Algorithm~\ref{alg:Elliptic:Input}; inputs are then solved using the FEM simulator described above.
Here, we investigate the impact of various modeling choices on the accuracy of the surrogate model when making forward predictions.
We consider between $n_\xi = 16$ and $1024$ input and output realizations.
We exploit the regularity of the spatial points at which the data are observed for our problem to define the structured matrix of spatial input points
\begin{equation}
    \mtx X^{(s)} = \mtx X^{(s, 1)} \otimes \mtx X^{(s, 2)},
\end{equation}
where $\mtx X^{(s, out, i)} \in \reals^{\tilde{n}_{s, out} \times 1}$ ($i = 1, 2$) contain the horizontal and vertical locations.
In all cases, we use $m_\xi = \textrm{min}(n_\xi/2, 128)$ stochastic inducing points and $d_\xi = \textrm{min}(n_{\xi, in} / 2, 128)$ latent dimensions.
We choose to work with $\log a_k$ as the input to our statistical model and $\hat u$ as the output.
Lastly, due to the fact that the inputs and outputs have very different scales, we found it was necessary to rescale the output data for the jointly-trained model so that the elements of $\mtx Y^{(out)}$ have unit variance.
The two-model approach does not require rescaling because the each submodel learns its own set of kernel hyperparameters.

For all SGPLVM models except the two-model output submodel, the latent variables are initialized by performing PCA on the training data
In all cases, the stochastic inducing points are initialized as a random subset of the rows of $\mtx X^{(\xi)}$.
An exponential kernel is used for each spatial dimension:
\begin{equation}
    k_{s, i_s}(\vc x_{i, i_s}^{(s)}, \vc x_{j, i_s}^{(s)}) = \exp \left(-\frac{\norm{x_{i, i_s} - x_{j, i_s}}}{l_{s, i_s}} \right), ~i_s = 1, \dots, d_s.
\end{equation}
For the stochastic kernel, three options were considered for the jointly-trained model and the input submodel of the two-model approach: an exponentiated quadratic (aka ``Gaussian'') kernel,
\begin{equation}
    k_{\xi, eq}(\vc x_{i, :}^{(\xi)}, \vc x_{j, :}^{(\xi)}) = \sigma_\xi^2 \exp \left[ -\frac{1}{2} \sum_{k=1}^{d_\xi} \left( 
        \frac{x_{ik}^{(\xi)} - x_{jk}^{*(\xi)}}{l_k}
    \right)^2 \right],
\end{equation}
a linear kernel,
\begin{equation}
    k_{\xi, lin}(\vc x_{i, :}^{(\xi)}, \vc x_{j, :}^{(\xi)}) = \sum_{k=1}^{d_\xi} \sigma_{\xi, k}^2 x_{ik}^{(\xi)} x_{jk}^{*(\xi)},
\end{equation}
and a sum kernel,
\begin{equation}
    k_{\xi, sum} = k_{\xi, lin} + k_{\xi, eq}.
\end{equation}
The output submodel of the two-model approach used an exponentiated quadratic kernel.

\begin{figure}[hbt]
    \centering
    \includegraphics[width=0.8\textwidth]{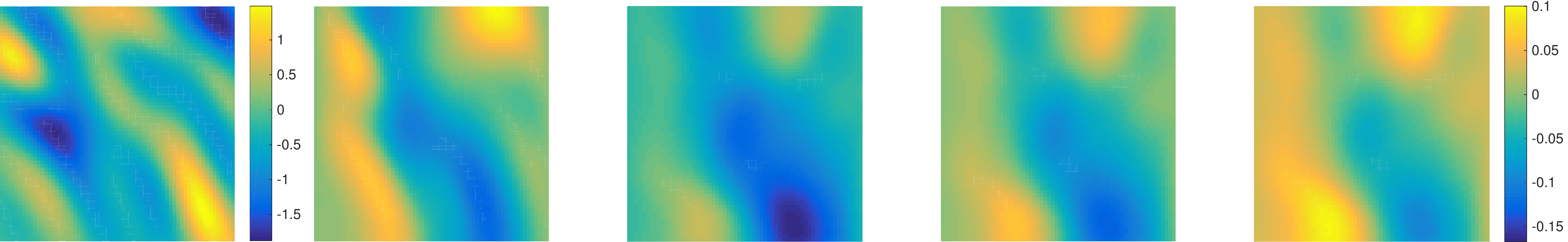}
    \\
    \includegraphics[width=0.8\textwidth]{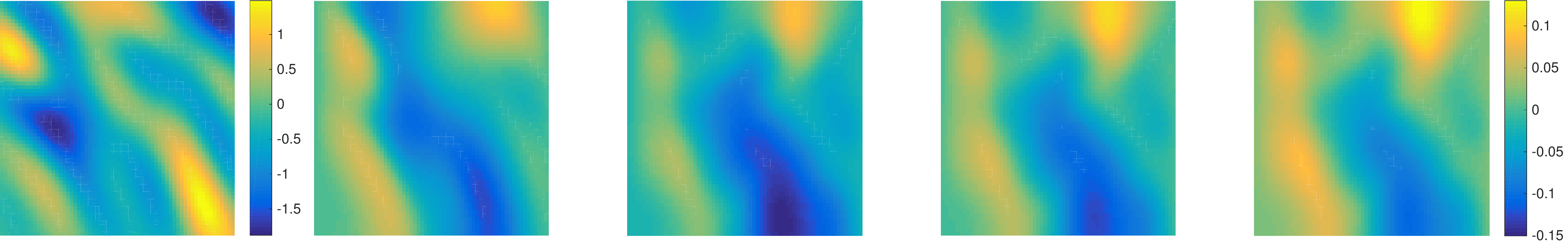}
    \\
    \includegraphics[width=0.8\textwidth]{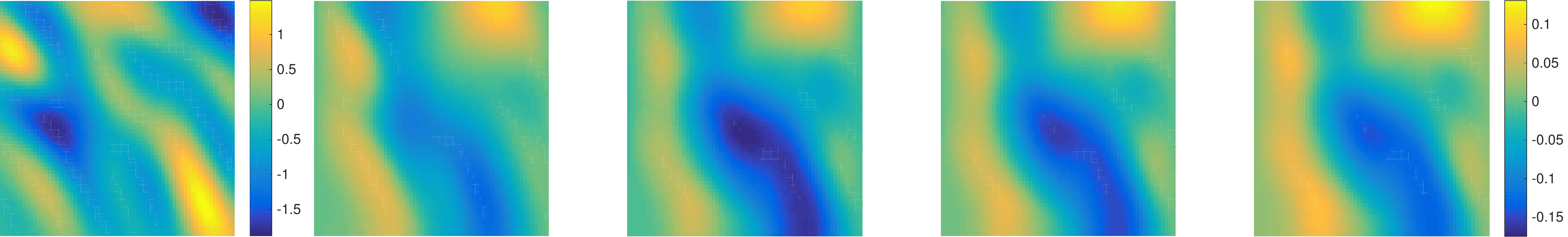}
    \\
    \includegraphics[width=0.8\textwidth]{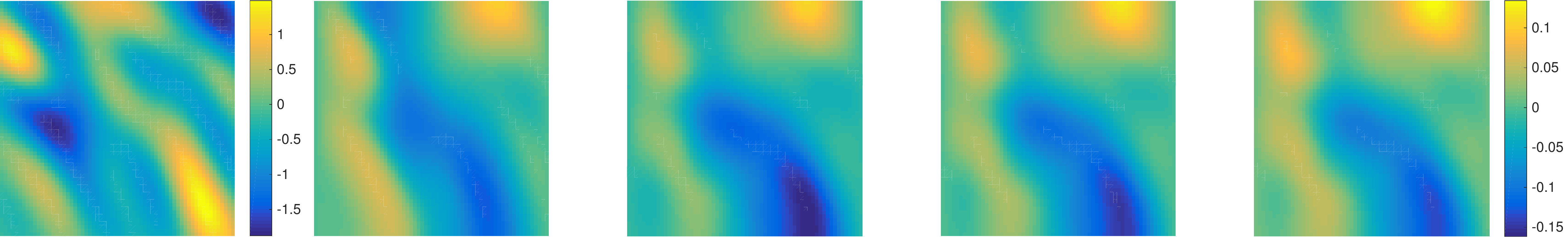}
    \\
    \includegraphics[width=0.8\textwidth]{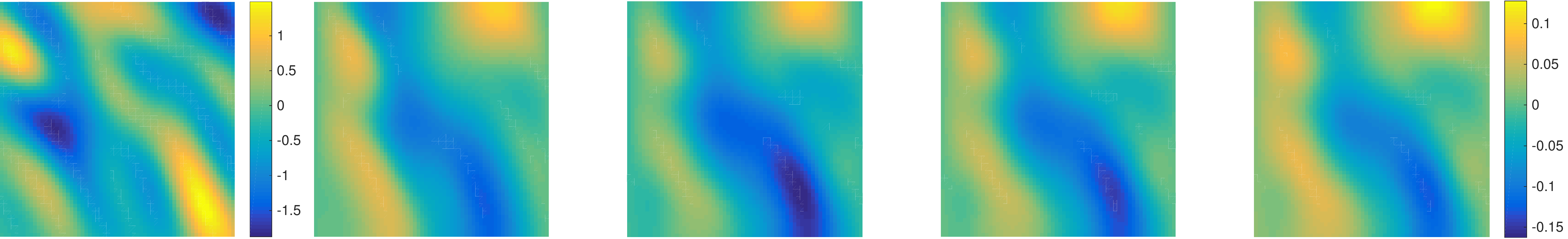}
    \\
    \includegraphics[width=0.8\textwidth]{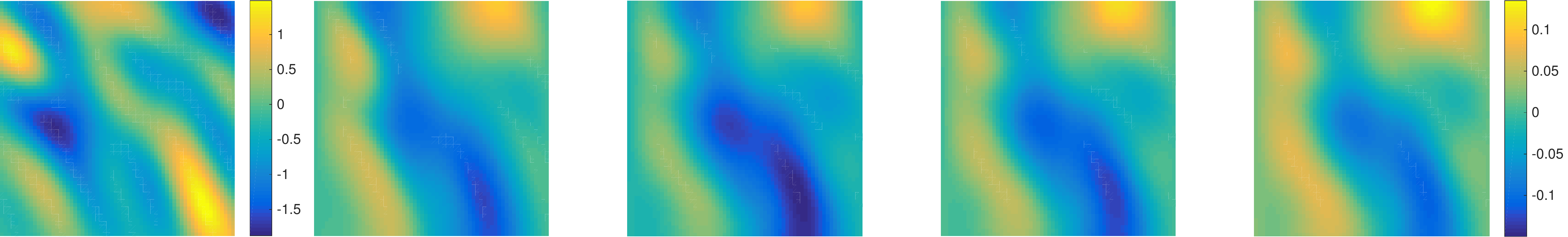}
    \caption{
        (Elliptic, forward, $\vc d_{KL}=(16, 32)$)
        Sample predictions with the 2M-Sum model.
        From top to bottom, $n_\xi = 32$, $64$, $128$, $256$, $512$, and $1024$.
        From left to right: Input, true output $\hat{u}$, prediction lower confidence bound, predictive mean, prediction upper confidence bound.
    }
    \label{fig:Elliptic:Forward:KL16-32:Predictions}
\end{figure}

\begin{figure}[hbt]
    \centering
     \includegraphics[width=0.8\textwidth]{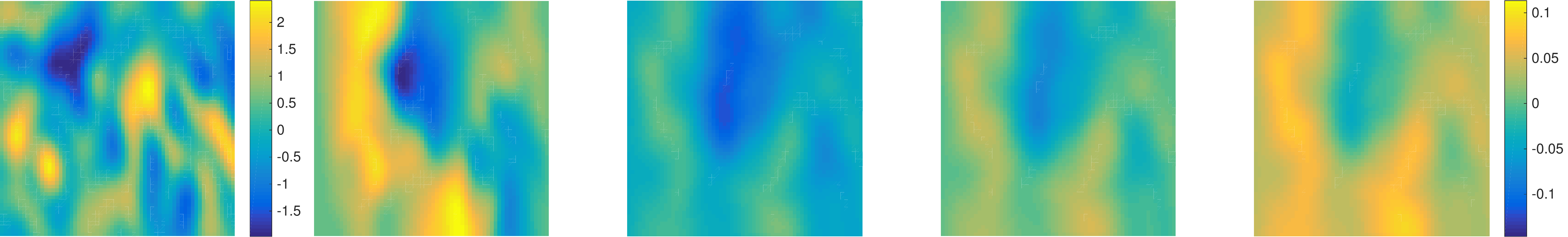}
    \\
    \includegraphics[width=0.8\textwidth]{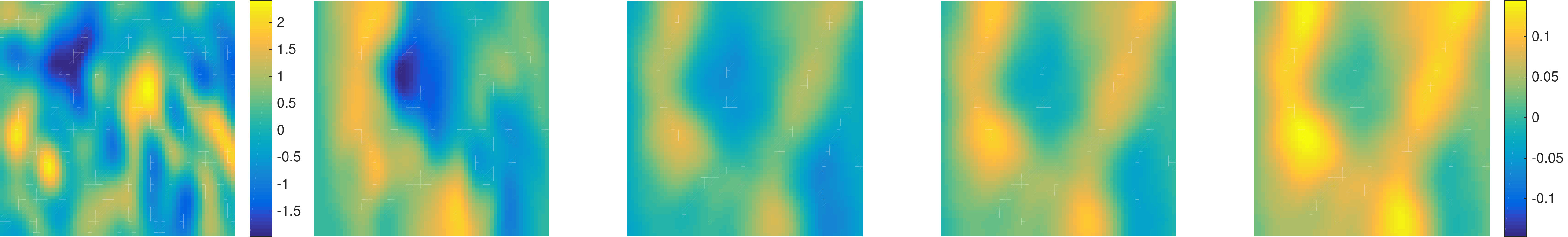}
    \\
    \includegraphics[width=0.8\textwidth]{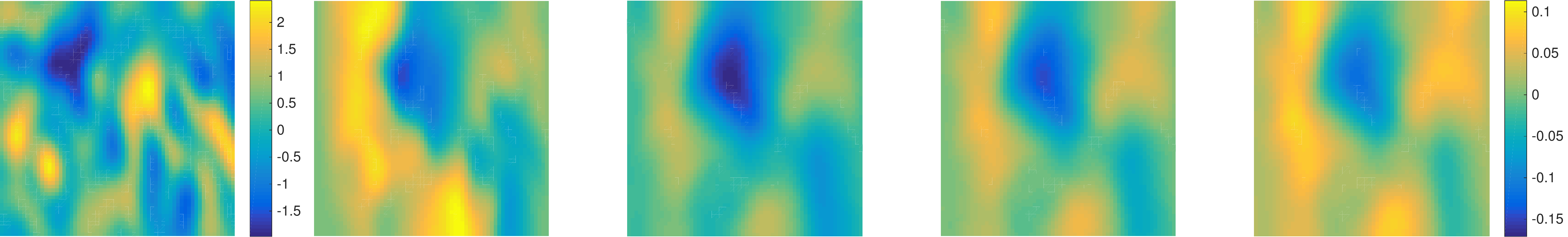}
    \\
    \includegraphics[width=0.8\textwidth]{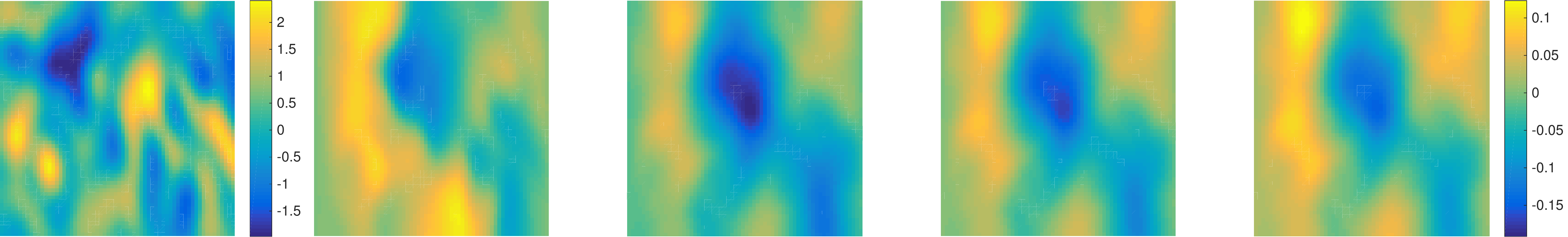}
    \\
    \includegraphics[width=0.8\textwidth]{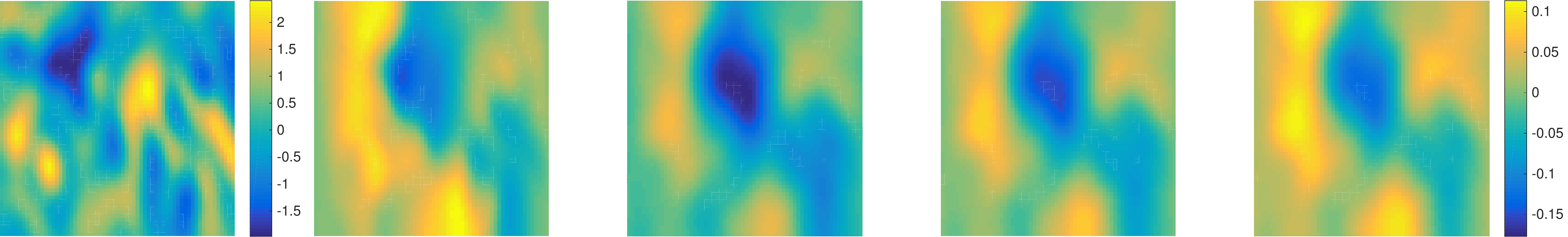}
    \\
    \includegraphics[width=0.8\textwidth]{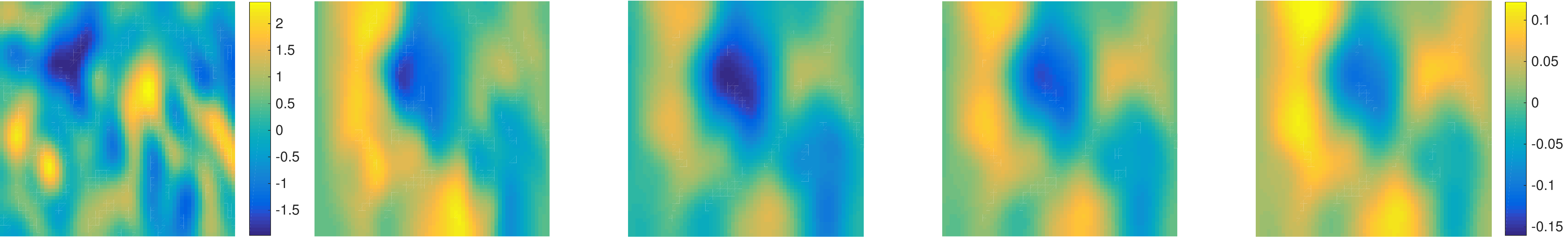}
    \caption{
        (Elliptic, forward, $\vc d_{KL}=(16, 128)$) 
        Sample predictions with the 2M-Sum model.
        The rows and columns are the same as in Fig.~\ref{fig:Elliptic:Forward:KL16-32:Predictions}.
    }   
    \label{fig:Elliptic:Forward:KL16-128:Predictions}
\end{figure}

We trained models using the two-model and jointly-trained model approaches following Algorithms~\ref{alg:Inverse:2Model:Train} and~\ref{alg:Inverse:JointModel:Train}.
To quantify the predictive accuracy of the trained models, predictions were carried out on $n_\xi^* = 100$ test realizations drawn from the stochastic prior model.
For each prediction, we measure the root mean squared error (RMSE),
\begin{equation}
    RMSE = \left( 
        \frac{1}{n_{s, out}^*} \sum_{i=1}^{n_{s, out}^*} \left(
            \bar{\mu}_i^{out, *} - y_i^{out, *}
        \right)^2
    \right)^{1/2},
\end{equation}
as well as the median of the log probability (MNLP), where the predictive density is approximated as a Gaussian with mean $\bar{\vc \mu}^*$ and variance $\bar{\mtx V}^*$ as described in Section~\ref{sec:Theory:SGPLVM:Predictions:Forward},
\begin{equation}
    MNLP = \textrm{median} \left\{
        \log \G{y_i^{out, *}}{\mu_i^{out, *}}{\bar{v}_i^{out, *}}
    \right\}_{i=1}^{n_{s, out}^*}.
\end{equation}
We report the prediction accuracy for the various models as a function of the number of training realizations $n_\xi$ in Tables~\ref{tab:Elliptic:Forward:KL16-32:RMSE}-\ref{tab:Elliptic:Forward:KL16-128:MNLP}.
Figures~\ref{fig:Elliptic:Forward:KL16-32:Predictions} and~\ref{fig:Elliptic:Forward:KL16-128:Predictions} illustrate a few example predictions with the trained models.

\begin{table}[htb]
    \center
    \begin{tabular}{|c|c|c|c|c|c|}
        \hline
        $n$ & PCA & 2M-Lin & 2M-RBF & 2M-Sum & JM-Sum \\
        \hline
        $32$   & $5.34$ ($1.97$) & $4.07$ ($1.54$) & $4.11$ ($1.46$) & $6.06$ ($3.68$) & $5.23$ ($2.12$) \\
        $64$   & $3.06$ ($1.30$) & $3.17$ ($1.39$) & $3.14$ ($1.37$) & $3.16$ ($1.30$) & $3.76$ ($1.51$) \\
        $128$  & $2.44$ ($1.00$) & $2.48$ ($0.96$) & $2.48$ ($0.94$) & $2.55$ ($0.99$) & $3.44$ ($1.48$) \\
        $256$  & $2.10$ ($1.08$) & $1.98$ ($0.97$) & $2.39$ ($1.07$) & $2.03$ ($0.97$) & $3.75$ ($1.59$) \\
        $512$  & $1.51$ ($0.69$) & $1.55$ ($0.72$) & $1.54$ ($0.71$) & $1.49$ ($0.67$) & $3.69$ ($1.53$) \\
        $1024$ & $1.23$ ($0.61$) & $1.22$ ($0.63$) & $1.24$ ($0.65$) & $1.25$ ($0.61$) & $3.62$ ($1.44$) \\
        \hline
    \end{tabular}
    \caption{
        (Elliptic, forward, $\vc d_{KL}=(16, 32)$) Mean RMSE $\times 10^2$ (and standard deviation) as a function of $n_\xi$.
        Columns correspond to using PCA for dimensionality reduction,
        the two-model approach with Linear, RBF, and linear+RBF stochastic kernels on the input model and RBF stochastic kernel on the output model,
        and 
        the jointly-trained approach with linear+RBF kernel.
    }
    \label{tab:Elliptic:Forward:KL16-32:RMSE}
\end{table}
\begin{table}[htb]
    \center
    \begin{tabular}{|c|c|c|c|c|c|}
         \hline
         $n$ & PCA & 2M-Lin & 2M-RBF & 2M-Sum & JM-Sum \\
         \hline
         $32$   & $-1.40$ ($2.64$) & $-1.65$ ($1.77$) & $-1.68$ ($1.61$) & $ 1.44$ ($7.57$) & $-1.73$ ($0.12$) \\
         $64$   & $-1.71$ ($2.02$) & $-1.50$ ($2.71$) & $-1.74$ ($2.19$) & $-1.60$ ($2.04$) & $-2.08$ ($0.21$) \\
         $128$  & $-1.23$ ($3.38$) & $-1.39$ ($2.55$) & $-1.08$ ($2.71$) & $-1.27$ ($2.63$) & $-1.82$ ($1.40$) \\
         $256$  & $-1.56$ ($3.34$) & $-1.73$ ($2.58$) & $-1.66$ ($2.38$) & $-1.59$ ($2.64$) & $ 3.18$ ($9.20$) \\
         $512$  & $-2.91$ ($1.38$) & $-2.87$ ($1.27$) & $-2.89$ ($1.47$) & $-3.09$ ($0.84$) & $-0.47$ ($3.95$) \\
         $1024$ & $-3.20$ ($1.16$) & $-3.30$ ($1.00$) & $-3.28$ ($0.99$) & $-3.34$ ($0.63$) & $-1.54$ ($1.98$) \\
         \hline
    \end{tabular}
    \caption{
        (Elliptic, forward, $\vc d_{KL}=(16, 32)$) Mean MNLP (standard deviation) as a function of $n_\xi$.
        Refer to Table~\ref{tab:Elliptic:Forward:KL16-32:RMSE} for details on the columns.
    }
    \label{tab:Elliptic:Forward:KL16-32:MNLP}
\end{table}

\begin{table}[htb]
    \center
   \begin{tabular}{|c|c|c|c|c|c|}
         \hline
         $n_\xi$ & PCA & 2M-Lin & 2M-RBF & 2M-Sum & JM-Sum \\
         \hline
         $32$   & $4.72$ ($1.39$) & $4.56$ ($1.39$) & $4.81$ ($1.59$) & $5.01$ ($1.64$) & $4.49$ ($2.86$) \\
         $64$   & $3.70$ ($1.16$) & $3.66$ ($1.11$) & $3.63$ ($1.07$) & $4.40$ ($1.92$) & $3.54$ ($2.33$) \\
         $128$  & $3.32$ ($1.04$) & $3.26$ ($1.04$) & $3.14$ ($1.01$) & $3.57$ ($1.30$) & $3.76$ ($2.15$) \\
         $256$  & $2.80$ ($1.02$) & $2.78$ ($0.98$) & $2.73$ ($0.98$) & $2.75$ ($0.97$) & $4.30$ ($1.42$) \\
         $512$  & $2.21$ ($0.67$) & $2.28$ ($0.69$) & $2.24$ ($0.70$) & $2.28$ ($0.66$) & $4.53$ ($1.57$) \\
         $1024$ & $2.79$ ($0.85$) & $1.91$ ($0.65$) & $2.87$ ($2.70$) & $1.86$ ($0.62$) & $4.94$ ($1.90$) \\
         \hline
    \end{tabular}
    \caption{
        (Elliptic, forward, $\vc d_{KL}=(16, 128)$) Mean RMSE $\times 10^2$ (standard deviation) as a function of $n_\xi$.
        Refer to Table~\ref{tab:Elliptic:Forward:KL16-32:RMSE} for details on the columns.
    }
    \label{tab:Elliptic:Forward:KL16-128:RMSE}
\end{table}

\begin{table}[htb]
    \center
    \begin{tabular}{|c|c|c|c|c|c|}
         \hline
         $n_\xi$ & PCA & 2M-Lin & 2M-RBF & 2M-Sum & JM-Sum \\
         \hline
         $32$   & $-1.62$ ($0.80$) & $-1.74$ ($1.05$) & $-1.53$ ($1.54$) & $-1.27$ ($1.77$) & $-1.43$ ($0.06$) \\
         $64$   & $-1.82$ ($1.46$) & $-2.01$ ($1.13$) & $-2.08$ ($1.08$) & $-1.28$ ($2.11$) & $-1.58$ ($0.04$) \\
         $128$  & $-1.66$ ($1.58$) & $-1.85$ ($1.42$) & $-2.00$ ($1.38$) & $-1.43$ ($2.25$) & $-1.83$ ($0.10$) \\
         $256$  & $-1.51$ ($2.28$) & $-1.79$ ($1.71$) & $-1.79$ ($1.93$) & $-1.86$ ($1.67$) & $-2.03$ ($0.31$) \\
         $512$  & $-2.71$ ($0.72$) & $-2.69$ ($0.71$) & $-2.70$ ($0.69$) & $-2.68$ ($0.71$) & $-1.93$ ($0.29$) \\
         $1024$ & $-2.33$ ($0.12$) & $-2.91$ ($0.48$) & $-2.69$ ($0.76$) & $-2.98$ ($0.41$) & $-1.87$ ($0.39$) \\
         \hline
    \end{tabular}
    \caption{
        (Elliptic, forward, $\vc d_{KL}=(16, 128)$) Mean MNLP (standard deviation) as a function of $n_\xi$.
        Refer to Table~\ref{tab:Elliptic:Forward:KL16-32:RMSE} for details on the columns.
    }
    \label{tab:Elliptic:Forward:KL16-128:MNLP}
\end{table}

We did not observe any consistent improvement by using the jointly-trained model.
This suggests that the latent space learned solely from the input data, combined with with the representative capacity of the output model, were adequate to capture the input-output relationship encoded by the data.
Furthermore, it seems that the inference step, in which a latent variable posterior is obtained from the test input, is the most challenging part of the prediction pipeline.
By incorporating information from the training outputs when constructing the latent space, we are making it more difficult to project to latent space at test time.

Finally, we compare our method to a simpler approach in which we perform data-driven dimensionality reduction using PCA and train a plain structured GP regression model to predict the outputs.
By doing so, we lose the ability to perform nonlinear dimensionality reduction and to propagate the epistemic uncertainty in the dimensionality reduction to the outputs. 
Indeed, we find that this approach performs worse than using an SGPLVM, particularly when the number of data is small.
This is consistent with the notion that the Bayesian formulation of the SGPLVM model allows us to accurately capture the epistemic uncertainty and guard against overfitting.

Next, we note that the sum kernel for input dimensionality reduction seems to perform most poorly when training data are limited, but becomes the best approach once $n_\xi$ becomes larger.
This may be because the sum kernel has more kernel hyperparameters to optimize, which can put the model in jeopardy of overfitting.
As the number of examples increase, this danger is ameliorated.

\subsection{Inverse problem}
\label{sec:Examples:Elliptic:Inverse}
In this section, we demonstrate use use of the variational method to invert partially-observed noisy data from the output space.
Here, we are provided with a set of noisy measurements $\mc D$ to the solution $u(\vc x_s)$ to Eq.\ (\ref{eqn:Elliptic}) on a set of spatial points $\mtx X^{(s, out, *)} \in \reals^{n_s^* \times d_s}$.
Given this data, we seek to infer the unknown field $a_k( \vc x_s)$ that produced this solution.

The models are defined and trained in the same way as in Section~\ref{sec:Examples:Elliptic:Forward}.
Predictions are carried out as described in Section~\ref{sec:Theory:Inverse}.
Importantly, when inferring the latent variable for each test case, we additionally optimize over $\beta_*$ of Eq.~(\ref{eqn:apx:SGPLVM:BoundTestTerm}) in order to account for the fact that the simulator data will not have the same level of noise as the test data, which are assumed to come from a different information source (such as a sensor array).

We consider two different experimental setups with increasing difficulty.
Again, we perform two experiments where training data is generated by solving samples from the stochastic prior with $\vc d_{KL}=(16, 32)$ in the first experiment and $(16, 128)$ in the second experiment,
Test cases are obtained by subsampling solutions to the FEM code on a $5 \times 5$ grid for the first experiment and a $9 \times 9$ grid for the second experiment.
For both experiments, observations are corrupted with white Gaussian noise with standard deviation roughly equal to 10\% of the standard deviation of a typical solution $\hat u$.

\begin{figure}[ht]
    \centering
    \includegraphics[width=0.8\textwidth]{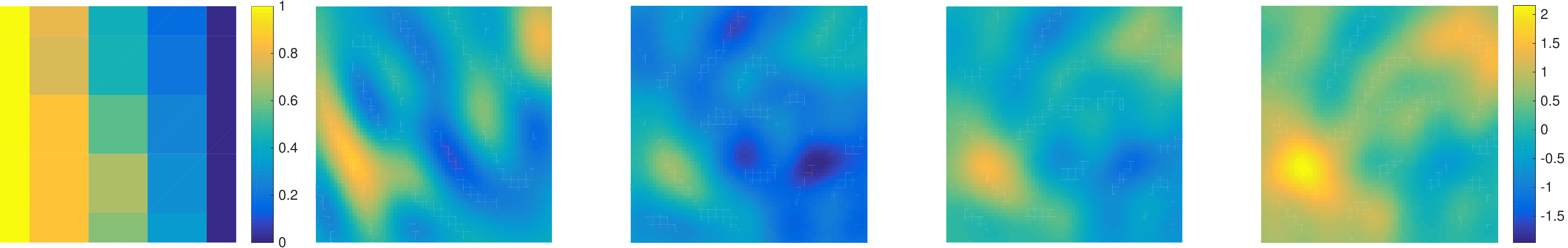}
    \\
    \includegraphics[width=0.8\textwidth]{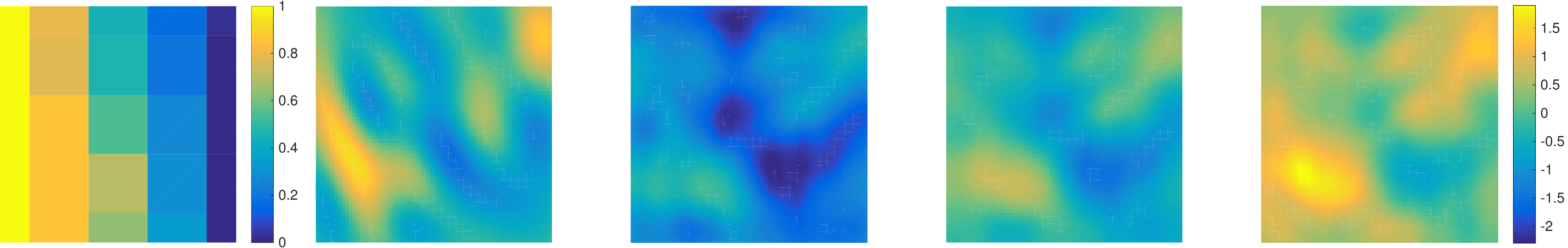}
    \\
    \includegraphics[width=0.8\textwidth]{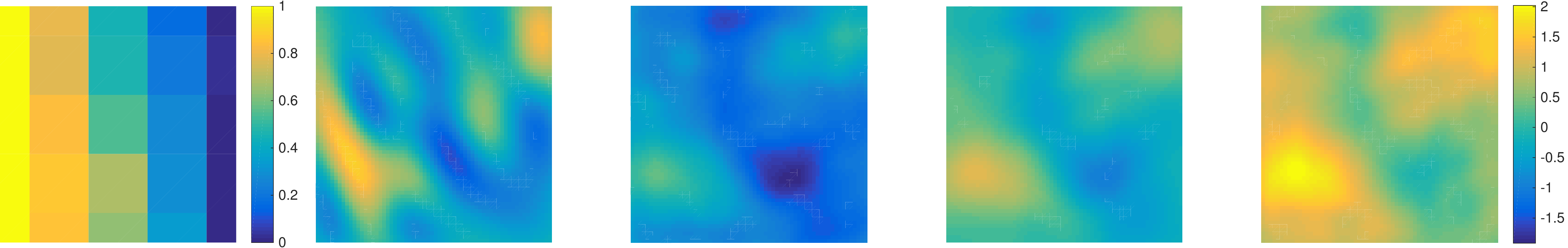}
    \\
    \includegraphics[width=0.8\textwidth]{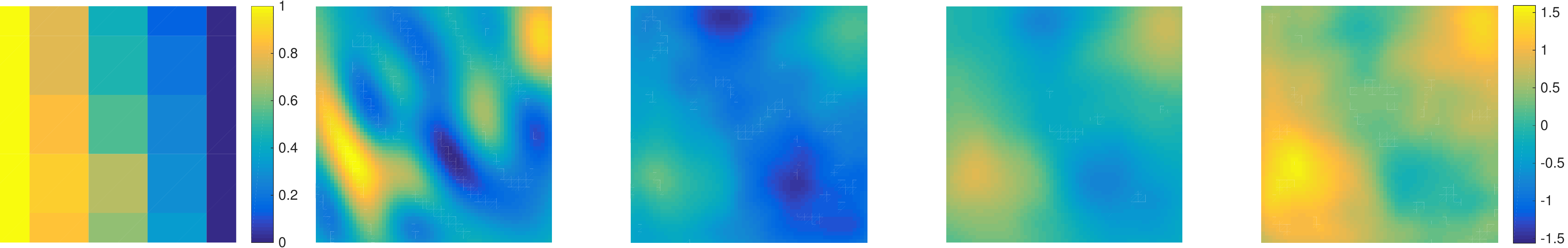}
    \\
    \includegraphics[width=0.8\textwidth]{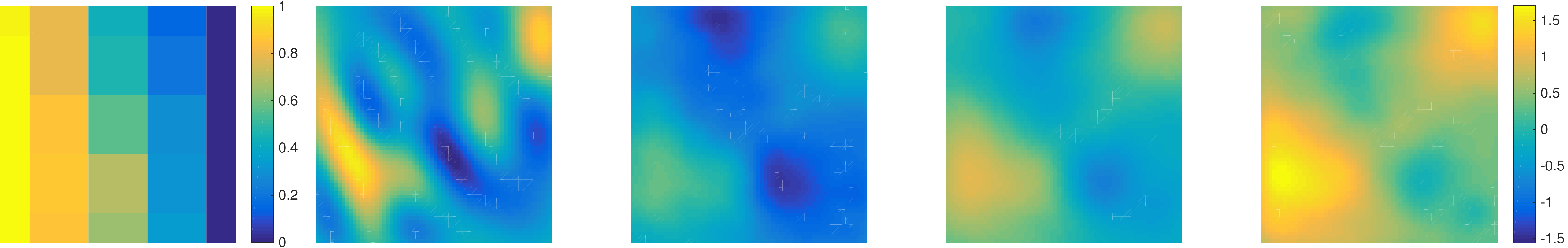}
    \\
    \includegraphics[width=0.8\textwidth]{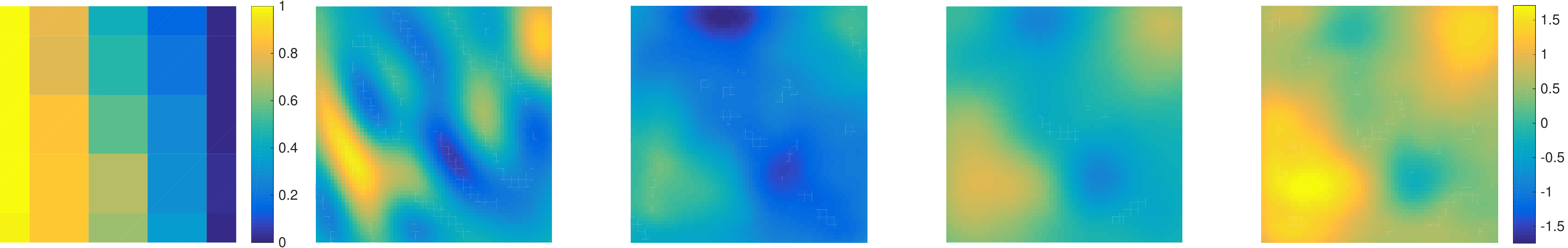}
    \caption{
        (Elliptic, inverse, $d_{KL}=(16, 32)$)
        Sample predictions from the JM-Sum model.
        From top to bottom, $n_\xi = 32$, $64$, $128$, $256$, $512$, and $1024$.
        From left to right: Measured output, true input $\log a_k$, prediction lower confidence bound, predictive mean, prediction upper confidence bound.
    }
    \label{fig:Elliptic:Inverse:KL32:2DExamples}
\end{figure}

\begin{figure}[ht]
    \centering
    \includegraphics[width=0.8\textwidth]{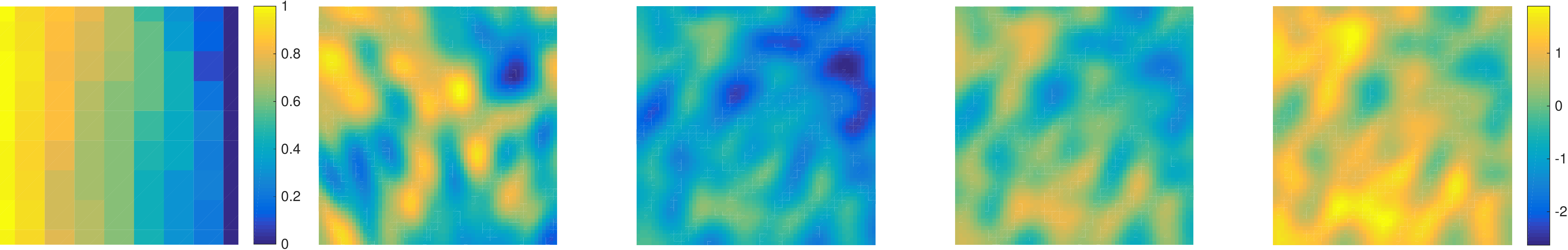}
    \\
    \includegraphics[width=0.8\textwidth]{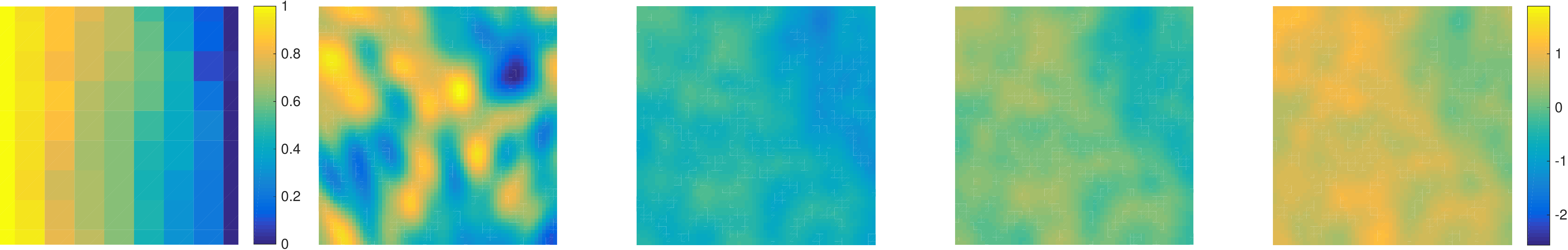}
    \\
    \includegraphics[width=0.8\textwidth]{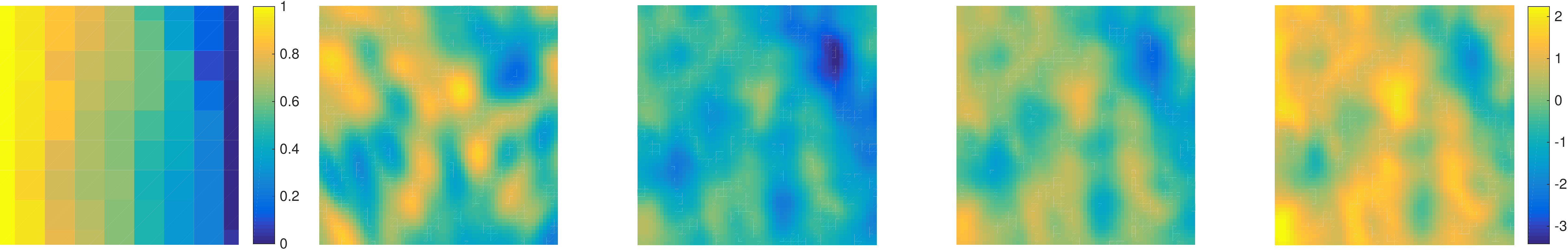}
    \\
    \includegraphics[width=0.8\textwidth]{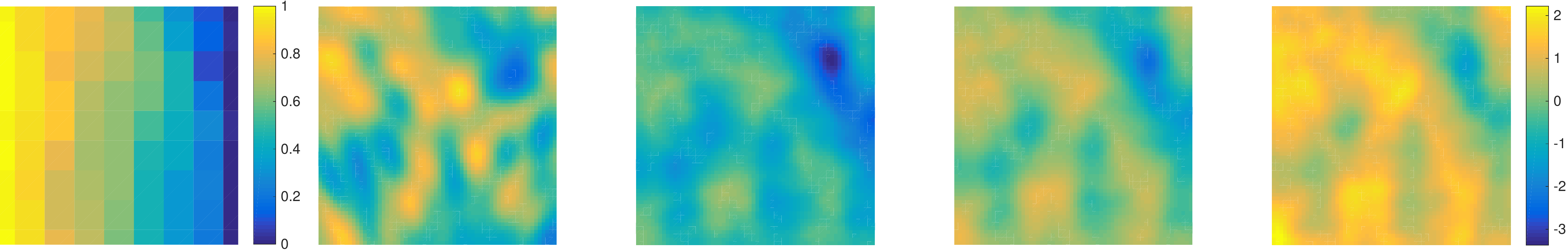}
    \\
    \includegraphics[width=0.8\textwidth]{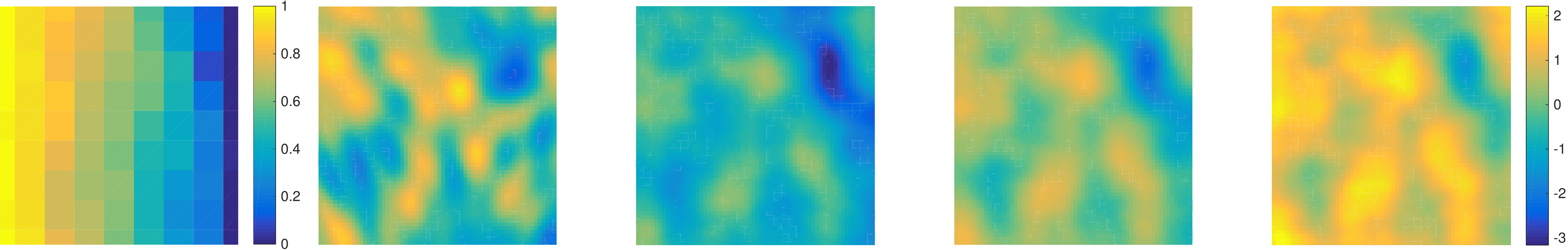}
    \\
    \includegraphics[width=0.8\textwidth]{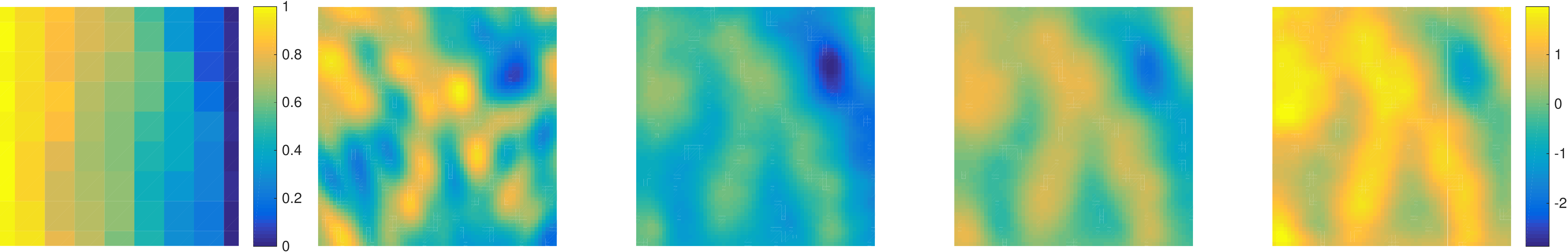}
    \caption{
        (Elliptic, inverse, $d_{KL}=(16, 128)$)
        Sample predictions from the JM-Sum model.
        From top to bottom, $n_\xi = 32$, $64$, $128$, $256$, $512$, and $1024$.
        From left to right: Measured output, true input $\log a_k$, prediction lower confidence bound, predictive mean, prediction upper confidence bound.
    }
    \label{fig:Elliptic:Inverse:KL128:2DExamples}
\end{figure}

Tables~\ref{tab:Elliptic:Inverse:KL32:RMSE} and~\ref{tab:Elliptic:Inverse:KL32:MNLP} quantifies the prediction accuracy of our approach on a test set of $n_\xi^* = 100$ separate examples for the first experimental setup.
The performance on the second experimental setup is reported in Tables~\ref{tab:Elliptic:Inverse:KL128:RMSE} and~\ref{tab:Elliptic:Inverse:KL128:MNLP}.
Figures~\ref{fig:Elliptic:Inverse:KL32:2DExamples} and~\ref{fig:Elliptic:Inverse:KL128:2DExamples} visualize an example test case for both experimental setups using models with two different numbers of training data.
We see that the predictive accuracy of the model improves as more training examples are provided and that the posteriors retain well-calibrated uncertainty estimates.
Furthermore, the joint model approach consistently outperforms the two-model approaches by a slight margin, implying that learning a shared latent space between inputs and outputs provides a slight advantage within the context of this problem, where the more challenging step of inferring the latent variable posterior is aided by the fact that the latent space was learned with the ability to generate the outputs explicitly incorporated through the joint training approach.
All approaches seem to approach the same level of performance as the number of training data becomes larger, implying that the limiting aspect of its performance becomes the quality and quantity of output measurements for the test cases.

\begin{figure}[ht]
    \centering
    \includegraphics[width=0.44\textwidth]{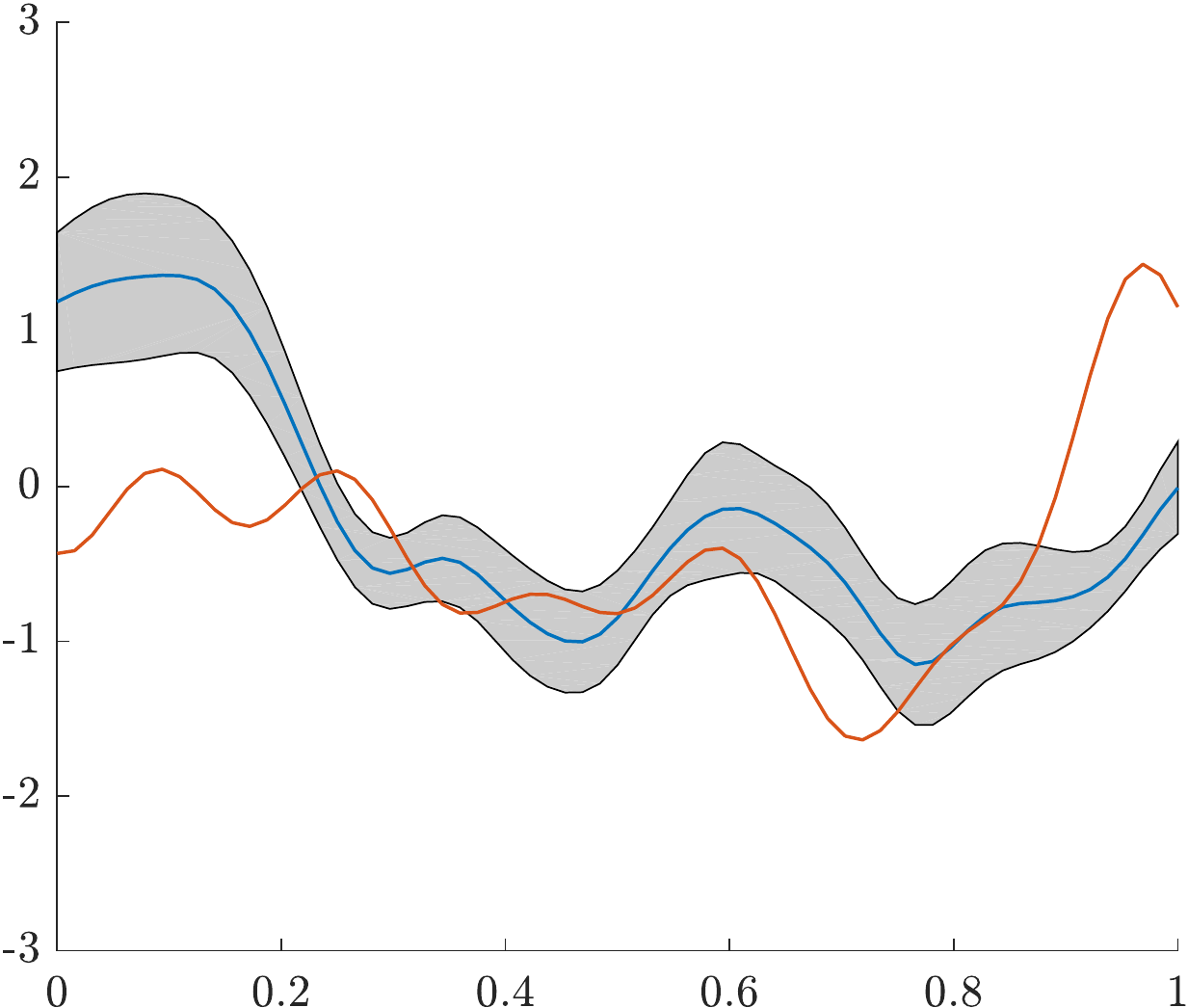}
    \includegraphics[width=0.44\textwidth]{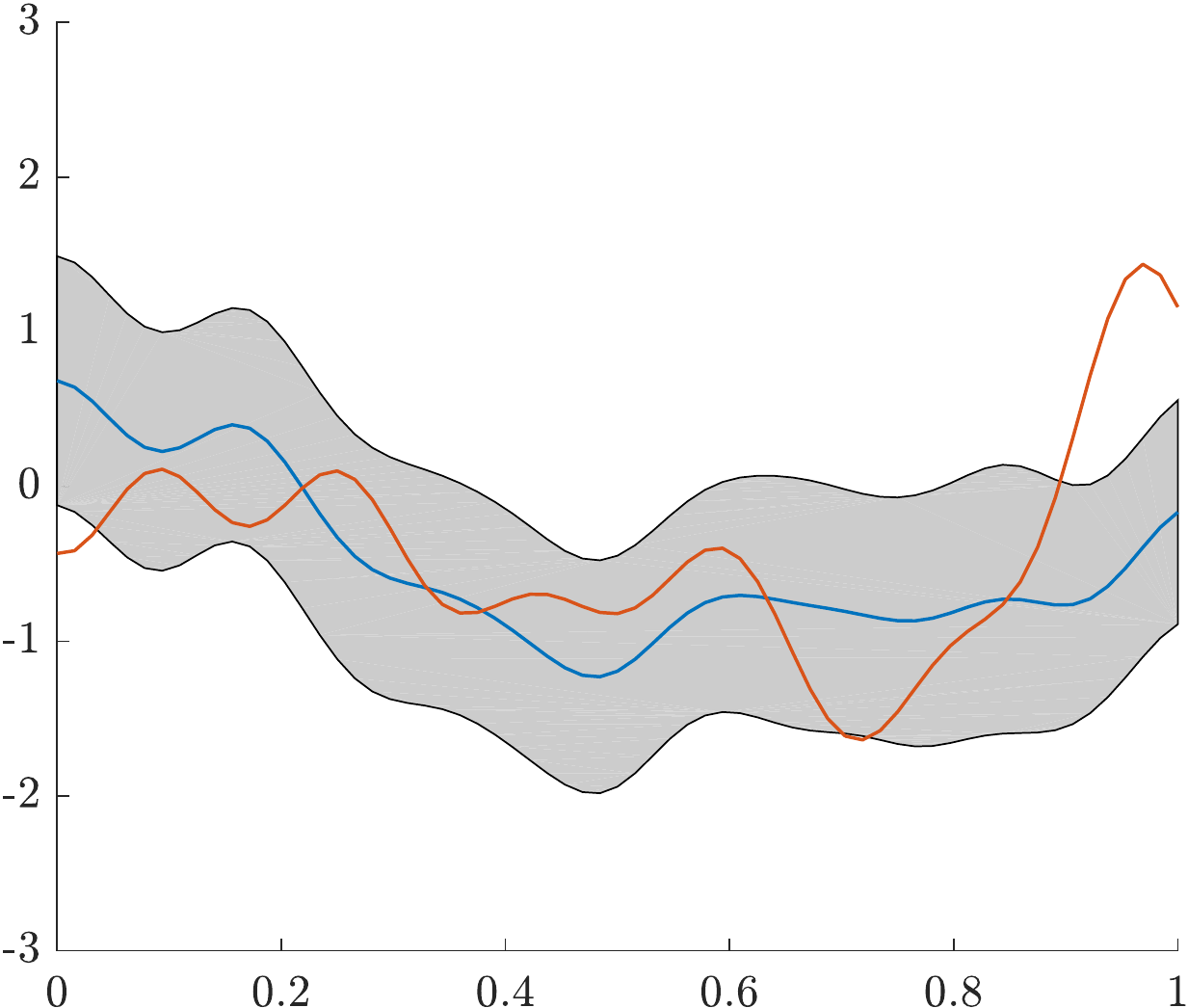}
    \caption{
        (Elliptic, inverse, $d_{KL}=(16, 128)$)
        One-dimensional cut along the line $x_1^{(s)} = 1/2$ of the posterior for a test case inverted using the (left) 2M-PCA and (right) JM-Sum models with $n_\xi = 32$.
    }
    \label{fig:Elliptic:Inverse:KL128:PcaCalibration}
\end{figure}

We again considered a case where deterministic PCA is used to construct the input model.
When $n_{\xi, in}$ is small, this approach suffers noticeably in its ability to provide well-calibrated posterior uncertainties as evidenced by the results in Tables~\ref{tab:Elliptic:Inverse:KL32:MNLP} and~\ref{tab:Elliptic:Inverse:KL128:MNLP}.
To a greater degree than in the forward problem, rigorously accounting for and propagating the sources of uncertainty in the model becomes imperative for obtaining good posterior estimates.
Thus, we see the benefits of the Bayesian formulation of the SGPLVM models.
To emphasize this point, Fig.~\ref{fig:Elliptic:Inverse:KL128:PcaCalibration} shows a one-dimensional cut of the posterior for one of the test cases in the second experiment along the line $x_1^{(s)} = 1/2$ for the 2M-PCA and JM-Sum models with $n_\xi = 32$.
We see that the former suffers from significant overconfidence, whereas the latter provides a well-calibrated posterior.

\begin{table}[htb]
    \center
    \begin{tabular}{|c|c|c|c|c|}
        \hline
        $n_{\xi, in}$ & $n_{\xi, out}$ & 2M-PCA & 2M-Sum & JM-Sum \\
        \hline
        $32$   & $32$   & $0.67$ ($0.12$) & $0.75$ ($0.16$) & $0.65$ ($0.13$) \\
        $64$   & $64$   & $0.60$ ($0.12$) & $0.60$ ($0.12$) & $0.60$ ($0.12$) \\
        $128$  & $128$  & $0.58$ ($0.11$) & $0.57$ ($0.11$) & $0.57$ ($0.10$) \\
        $256$  & $256$  & $0.58$ ($0.12$) & $0.56$ ($0.11$) & $0.58$ ($0.12$) \\
        $512$  & $512$  & $0.56$ ($0.11$) & $0.55$ ($0.11$) & $0.57$ ($0.12$) \\
        $1024$ & $1024$ & $0.55$ ($0.11$) & $0.60$ ($0.11$) & $0.56$ ($0.11$) \\
        \hline
    \end{tabular}
    \caption{
        (Elliptic, inverse, $d_{KL}=(16, 32)$) Mean RMSE (standard deviation) as a function of the number of training examples.
        Columns correspond to 
        (2M-PCA) two-model approach with a PCA input model and SGPLVM output model with linear+RBF sum kernel,
        (2M-Sum) the two-model approach with a linear+RBF stochastic kernel on both input submodels,
        and 
        (JM-Sum) the jointly-trained approach with linear+RBF kernel.
    }
    \label{tab:Elliptic:Inverse:KL32:RMSE}
\end{table}

\begin{table}[htb]
    \center
    \begin{tabular}{|c|c|c|c|c|}
        \hline
        $n_{\xi, in}$ & $n_{\xi, out}$ & 2M-PCA & 2M-Sum & JM-Sum \\
        \hline
        $32$   & $32$   & $1.70$ ($1.45$) & $1.01$ ($0.54$) & $0.82$ ($0.50$) \\
        $64$   & $64$   & $0.60$ ($0.33$) & $0.57$ ($0.25$) & $0.60$ ($0.27$) \\
        $128$  & $128$  & $0.53$ ($0.19$) & $0.55$ ($0.20$) & $0.51$ ($0.20$) \\
        $256$  & $256$  & $0.54$ ($0.21$) & $0.50$ ($0.38$) & $0.61$ ($0.36$) \\
        $512$  & $512$  & $0.50$ ($0.19$) & $0.52$ ($0.17$) & $0.55$ ($0.32$) \\
        $1024$ & $1024$ & $0.49$ ($0.21$) & $0.64$ ($0.41$) & $0.51$ ($0.28$) \\
        \hline
    \end{tabular}
    \caption{
        (Elliptic, inverse, $d_{KL}=(16, 32)$) Mean MNLP (standard deviation) as a function of the number of training examples.
        Refer to Table~\ref{tab:Elliptic:Inverse:KL32:RMSE} for details on the columns.
    }
    \label{tab:Elliptic:Inverse:KL32:MNLP}
\end{table}

\begin{table}[htb]
    \center
    \begin{tabular}{|c|c|c|c|c|}
        \hline
        $n_{\xi, in}$ & $n_{\xi, out}$ & 2M-PCA & 2M-Sum & JM-Sum \\
        \hline
        $32$   & $32$   & $0.90$ ($0.13$) & $0.92$ ($0.14$) & \textbf{$0.88$ ($0.13$)} \\
        $64$   & $64$   & $0.82$ ($0.12$) & $0.86$ ($0.12$) & $0.78$ ($0.10$) \\
        $128$  & $128$  & $0.76$ ($0.12$) & $0.75$ ($0.12$) & $0.72$ ($0.12$) \\
        $256$  & $256$  & $0.68$ ($0.10$) & $0.68$ ($0.10$) & $0.67$ ($0.10$) \\
        $512$  & $512$  & $0.66$ ($0.10$) & $0.75$ ($0.13$) & $0.65$ ($0.10$) \\
        $1024$ & $1024$ & $0.64$ ($0.10$) & $0.66$ ($0.10$) & $0.65$ ($0.10$) \\
        \hline
    \end{tabular}
    \caption{
        (Elliptic, inverse, $d_{KL}=(16, 128)$) Mean RMSE (standard deviation) as a function of the number of training examples.
        Refer to Table~\ref{tab:Elliptic:Inverse:KL32:RMSE} for details on the columns.
    }
    \label{tab:Elliptic:Inverse:KL128:RMSE}
\end{table}

\begin{table}[htb]
    \center
    \begin{tabular}{|c|c|c|c|c|}
        \hline
        $n_{\xi, in}$ & $n_{\xi, out}$ & 2M-PCA & 2M-Sum & JM-Sum \\
        \hline
        $32$   & $32$   & $5.88$ ($2.78$) & $1.11$ ($0.42$) & $1.13$ ($0.35$) \\
        $64$   & $64$   & $3.11$ ($1.66$) & $0.98$ ($0.37$) & $1.30$ ($0.52$) \\
        $128$  & $128$  & $1.20$ ($0.57$) & $0.79$ ($0.34$) & $0.84$ ($0.40$) \\
        $256$  & $256$  & $0.69$ ($0.27$) & $0.66$ ($0.21$) & $0.69$ ($0.28$) \\
        $512$  & $512$  & $0.66$ ($0.23$) & $0.84$ ($0.31$) & $0.65$ ($0.22$) \\
        $1024$ & $1024$ & $0.64$ ($0.24$) & $0.67$ ($0.26$) & $0.69$ ($0.29$) \\
        \hline
    \end{tabular}
    \caption{
         (Elliptic, inverse, $d_{KL}=(16, 128)$) Mean MNLP (standard deviation) as a function of the number of training examples.
        Refer to Table~\ref{tab:Elliptic:Inverse:KL32:RMSE} for details on the columns.
    }
    \label{tab:Elliptic:Inverse:KL128:MNLP}
\end{table}

\section{Conclusions and Discussion}
\label{sec:Conclusion}
In this work, we derived a structured Gaussian process latent variable model that can model spatiotemporal data, explicitly capturing spatiotemporal correlations by extending the Bayesian GP-LVM of Titsias and Lawrence.
Computational tractability is maintained by expressing the prohibitively-large covariance matrices involved in the model's evidence lower bound and predictive density as Kronecker products.
Additionally, the modeled spatiotemporal correlations are expressed in terms of standard kernel functions, yielding a simple, interpretable parameterization and allowing for one to use the generative model at a higher resolution than that of the training data.
Finally, we use the SGPLVM to solve forward and inverse problems associated with an elliptic PDE, including showing how the inverse problem can be solved through variational optimization and avoiding costly Monte Carlo sampling.

Our work might be extended in a number of ways.
First, it is possible to express the latent variables as outputs of a second GP, resulting in a structured version of the Bayesian warped Gaussian process first found in~\cite{lazaro2012bayesian}.
Such a model might prove useful as a surrogate model in which the data exhibits nonstationary behavior in the stochastic inputs.
Taking this even further, one could append additional layers or other architectures to map from some inputs to the latent variables found in the SGPLVM; such a model would be tractable using techniques developed for deep Gaussian processes~\cite{damianou2013deep} or deep kernel learning~\cite{wilson2015kernel}, to name two possibilities.
Additionally, one could perform subsequent manipulations on the output side of the structured GP of the current model; such a model would be useful for learning nonparametric, non-Gaussian likelihoods and improve the applicability of the current generative model to binary (or strongly multimodal) data.

Lastly, the use of variational methods and suitably-posed probabilistic generative models for Bayesian inversion is attractive to us particularly for applications where speed is essential.
Along this same line, we are investigating other methods utilizing parametric deep learning models including generative adversarial networks.

\section*{Acknowledgements}
The work is supported by the Computational Mathematics Program of AFOSR. 
The computing 
was facilitated by the resources of the University of Notre Dame's Center for 
Research Computing (CRC). Additional computing resources were provided by the  NSF supported ``Extreme Science and Engineering Discovery Environment'' (XSEDE) on the Bridges and Bridges-GPU cluster  
through the startup allocation No.  TG-DMS$180011$.

\clearpage
\appendix

\section{SGPR: Predictive covariance computation}
\label{apx:SGPR:PredictiveCovariance}
We wish to compute the predictive covariance of Eq.~(\ref{eqn:Theory:GPR:Predictions}) for the structured GP model, repeated here for convenience:
\begin{equation}
    \vc \Sigma^* = \mtx K_{**} - \ksf \kyy^{-1} \kfs.
    \nonumber
\end{equation}
We will restrict ourselves to the case where $n_\xi^* = 1$, since larger values will typically outpace our ability to store $\mtx \Sigma^*$ in memory.
If one wishes to sample a larger distribution, various sequential/low-rank approaches such as the one used in~\cite{bilionis2013multi} might be adapted to the SGPR model.

Using Eq.\ (\ref{eqn:SGPR:KyyInverse}), we can write
\begin{equation}
    \ksf \kyy^{-1} \kfs = \ksf \mtx Q (\mtx \Lambda + \beta^{-1} \mtx I_{n \times n})^{-1} \mtx Q^\intercal \kfs.
\end{equation}

Define
\begin{align}
    \tilde{\mtx D}_f &= (\mtx \Lambda + \beta^{-1} \mtx I_{n \times n})^{-1/2},
    \\
    \mtx E_f &= \ksf \mtx Q.
\end{align}
Also, define $\tilde{\mtx D}_f^{(i)} \in \reals^{n_s \times n_s}$ to be the $i$-th block diagonal of $\tilde{\mtx D}_f$.
The matrix $\mtx E_f$ has Kronecker structure:
\begin{align}
    \mtx E_f &= (\ksf^{(\xi)} \otimes \ksf^{(s)}) (\mtx Q^{(\xi)} \otimes \mtx Q^{(s)}) 
    \nonumber
    \\
    &= (\ksf^{(\xi)} \mtx Q^{(\xi)}) 
    \otimes
    (\ksf^{(s)} \mtx Q^{(s)}) 
    \nonumber
    \\
    &= \mtx E_f^{(\xi)} \otimes \mtx E_f^{(s)}.
\end{align}
Note that, in the case where we are interested in repeatedly predicting at the same $\xsts$, we can pre-compute $\mtx E_f^{(s)}$.
We can also pre-compute $\mtx Q^{(\xi)}$ and $\tilde{\mtx D}_f$ since they depend only on the training data.
The algorithm for computing $\mtx \Sigma^*$ for the SGPR model is given in Algorithm~\ref{alg:SGPR:PredictiveCovariance}.

\begin{algorithm}
    \textbf{Require:} A trained SGPR model, $\xsts$, $\vc x^{(\xi), *}$, pre-computed $\mtx E_f^{(s)}$, $\mtx Q^{(\xi)}$, and $\tilde{\mtx D}_f$.
    
    \textbf{Ensure:} The predictive covariance $\mtx \Sigma^*$ of Eq.\ (\ref{eqn:Theory:GPR:Predictions}).
    
    \begin{algorithmic}[1]
        \State Compute $\mtx E_f^{(\xi)} = \ksf^{(\xi)} \mtx Q^{(\xi)}$.
        \State Initialize $\mtx \Sigma^* \rightarrow 0$
        \For{$i=1, \dots, m_{\xi}$}
            \State Compute $\mtx H_f = \mtx E_f^{(s)} \tilde{\mtx D}_f^{(i)}$.
            \State Update $\mtx \Sigma^* \rightarrow \mtx \Sigma^* - (\mtx E_f^{(\xi)})_i \mtx H_f \mtx H_f^\intercal$.
        \EndFor
        \State Update $\mtx \Sigma^* \rightarrow \mtx \Sigma^* + \kss$.
    \end{algorithmic}
    \caption{Computing the predictive covariance for the SGPR model.}
    \label{alg:SGPR:PredictiveCovariance}
\end{algorithm}

\section{SGPLVM: Predictive covariance computation}
\label{apx:SGPLVM:PredictiveCovariance}
We wish to compute Eq.~(\ref{eqn:Theory:SGPLVM:x_to_y_covariance_sub}), repeated here for convenience:
\begin{equation}
    \mtx \Sigma^* = \mtx K_{**} 
    - \ksu \mtx L^{-\intercal} \mtx Q_C \left(\mtx I - \beta^{-1} \mtx D^{-1} \right) \mtx Q_C^\intercal \mtx L^{-1}
    \kus.
    \nonumber
\end{equation}
We will restrict ourselves to the case where $n_\xi^* = 1$, since larger values will typically outpace our ability to store $\mtx \Sigma^*$ in memory.
If one wishes to sample a larger distribution, various sequential/low-rank approaches such as the one used in~\cite{bilionis2013multi} might be adapted to the SGPLVM model.

Our approach is similar to that for the SGPR model.
Define
\begin{align}
    \tilde{\mtx D} &= (\mtx I_{m \times m} - \beta^{-1} \mtx D^{-1})^{1/2},
    \\
    \mtx E &= \mtx L^{-\intercal} \mtx Q_C,
    \\
    \mtx G &= \ksu \mtx L^{-\intercal} \mtx Q_C = \ksu \mtx E.
\end{align}
Also, define $\tilde{\mtx D}^{(i)} \in \reals^{n_{s} \times n_{s}}$ to be the $i$-th block diagonal of $\tilde{\mtx D}$.
The quantities $\mtx E$ and $\mtx G$ have Kronecker structure:
\begin{align}
    \mtx E &= \mtx E^{(\xi)} \otimes \mtx E^{(s)},
    \\
    \mtx G &= \mtx G^{(\xi)} \otimes \mtx G^{(s)}.
\end{align}
Note that, in the case where we are interested in repeatedly predicting at the same $\xsts$, we can pre-compute $\mtx G^{(s)}$.
We can also pre-compute $\mtx E^{(\xi)}$ since it depends only on the training data.
The algorithm for computing $\mtx \Sigma^*$ is given in Algorithm~\ref{alg:SGPLVM:PredictiveCovariance}.

\begin{algorithm}
    \textbf{Require:} A trained SGPLVM, $\xsts$, $\vc x^{(\xi), *}$, pre-computed $\mtx G^{(s)}$ $\mtx E^{(\xi)}$.
    
    \textbf{Ensure:} The predictive covariance $\mtx \Sigma^*$ of Eq.\ (\ref{eqn:Theory:SGPLVM:x_to_y_covariance_sub}).
    
    \begin{algorithmic}[1]
        \State Compute $\mtx G^{(\xi)} = \ksu^{(\xi)} \mtx E^{(\xi)}$.
        \State Initialize $\mtx \Sigma^* \rightarrow 0$
        \For{$i=1, \dots, m_{\xi}$}
            \State Compute $\mtx H = \mtx G^{(s)} \tilde{\mtx D}^{(i)}$.
            \State Update $\mtx \Sigma^* \rightarrow \mtx \Sigma^* - g_i^{(\xi)} \mtx H \mtx H^\intercal$.
        \EndFor
        \State Update $\mtx \Sigma^* \rightarrow \mtx \Sigma^* + \kss$.
    \end{algorithmic}
    \caption{Computing the predictive covariance.}
    \label{alg:SGPLVM:PredictiveCovariance}
\end{algorithm}

\section{Efficient computation of $\mc L^*$}
\label{apx:SGPLVM:TestInference}
Here, we provide details on the efficient computation of the test term $\mc L^*$ in the augmented bound.
This was first reported in~\cite{atkinson2018fully}, but is repeated here for convenience.
We are interested in computing
\begin{equation}
    \begin{aligned}
        \mc L^* = &-\frac{n^* d_y}{2} \left( \log(2 \pi) - \log \beta \right) 
        - \frac{\beta d_y}{2} \trace{\mtx Y^* \mtx Y^{*, \intercal}}
        - \frac{\beta}{2} \trace{
            \kuu^{-1} \psitwo^* \kuu^{-1} (\Ubar \Ubar^\intercal + d_y \mtx \Sigma_u)
        }
        \\
        &+ \beta \trace{\mtx Y^{*, \intercal} \psione^* \kuuinv \Ubar}
        - \frac{\beta d_y}{2} \left( \psi_0^* - \trace{\kuu^{-1} \psitwo^*} \right)
        - \KL{q(\vc x^{(\xi, *)})}{p(\vc x^{(\xi, *)})}.
    \end{aligned}
    \label{eqn:apx:SGPLVM:BoundTestTerm}
\end{equation}
The Kullback-Liebler term presents no particular challenges, so we will focus on computing $\mc F^* = \mc L^* + \KL{q(\vc x^{(\xi, *)})}{p(\vc x^{(\xi, *)})}$.
Recall that the analytic optimal variational posterior mean and covariance for $q^*(\mtx U) = \prod_{j=1}^{d_y} \G{\vc u_{:, j}}{\ubar_{:, j}^*}{\mtx \Sigma_u^*}$, given the training data, are given by
\begin{equation}
    \begin{aligned}
        \Ubar^* &= \kuu \kpsiinv \psione^\intercal \mtx Y,
        \\
        \mtx \Sigma_u^* &= \beta^{-1} \kuu \kpsiinv \kuu.
    \end{aligned}
    \label{eqn:apx:SGPLVM:OptimalU}
\end{equation}
Substituting these into Eq.~(\ref{eqn:apx:SGPLVM:BoundTestTerm})  and cleaning up some terms gives
\begin{equation}
    \begin{aligned}
        \mc F^* = &-\frac{n^* d_y}{2} \left( \log(2 \pi) - \log \beta \right) 
        - \frac{\beta  d_y}{2} \trace{\mtx Y^* \mtx Y^{*, \intercal}}
        \\
        &- \frac{\beta}{2} \trace{
            \psitwo^* 
            \kpsiinv \psione^\intercal \mtx Y \mtx Y^\intercal \psione \kpsiinv
        }
        - \frac{d_y}{2} \trace{
            \psitwo^* \kpsiinv
        }
        \\
        &+ \beta \trace{
            \mtx Y^{*, \intercal} \psione^*  \kpsiinv \psione^\intercal \mtx Y
        }
        - \frac{\beta d_y}{2} \left( \psi_0^* - \trace{\kuu^{-1} \psitwo^*} \right).
    \end{aligned}
    \label{eqn:apx:SGPLVM:BoundTestTerm2}
\end{equation}
Next, we compute the matrix square root
\begin{equation}
    \psitwo^* = 
    \underbrace{\left(\mtx L_{\psitwo}^{(\xi, *)} \otimes \kus^{(s)} \right)}_{= \mtx L_{\psitwo}^*}
    \left(\mtx L_{\psitwo}^{(\xi, *)} \otimes \kus^{(s)} \right)^\intercal,
\end{equation}
and substitute
\begin{equation}
    \kpsiinv = \mtx L^{-\intercal} \mtx Q_C \mtx D^{-1} \mtx Q_C^\intercal \mtx L^{-1}
    \label{eqn:apx:SGPLVM:KPsiInverse}
\end{equation}
into Eq.~(\ref{eqn:apx:SGPLVM:BoundTestTerm2}) to obtain
\begin{align}
    \begin{aligned}
        \mc F^* = &-\frac{n^* d_y}{2} \left( \log(2 \pi) - \log \beta \right) 
        - \frac{\beta  d_y}{2} \trace{\mtx Y^* \mtx Y^{*, \intercal}}
        \\
        &- \frac{\beta}{2} \trace{
            \underbrace{\mtx L_{\psitwo}^{*, \intercal}
            \mtx L^{-\intercal} \mtx Q_C \mtx D^{-1} \mtx Q_C^\intercal \mtx L^{-1}
            \psione^\intercal \mtx Y}_{= \mtx T_1}
            \mtx Y^\intercal \psione 
            \mtx L^{-\intercal} \mtx Q_C \mtx D^{-1} \mtx Q_C^\intercal \mtx L^{-1}
            \mtx L_{\psitwo}^*
        }
        \\
        &- \frac{d_y}{2} \trace{
            \underbrace{\mtx L_{\psitwo}^{*, \intercal}
            \mtx L^{-\intercal} \mtx Q_C \mtx D^{-1/2}}_{= \mtx T_2}
            \mtx D^{-1/2} \mtx Q_C^\intercal \mtx L^{-1}
            \mtx L_{\psitwo}^*
        }
        \\
        &+ \beta \trace{
            \mtx Y^{*, \intercal} \psione^* \kpsiinv \psione^\intercal \mtx Y
        }
        - \frac{\beta d_y}{2} \left( \psi_0^* - \trace{\kuu^{-1} \psitwo^*} \right),
    \end{aligned}
    \label{eqn:apx:SGPLVM:BoundTestTerm3}
    \\
    \begin{aligned}
        =  &-\frac{n^* d_y}{2} \left( \log(2 \pi) - \log \beta \right) 
        - \frac{\beta  d_y}{2} \trace{\mtx Y^* \mtx Y^{*, \intercal}}
        - \frac{\beta}{2} \trace{\mtx T_1 \mtx T_1^\intercal
        }
        - \frac{d_y}{2} \trace{\mtx T_2 \mtx T_2^\intercal
        }
        \\
        &+ \beta \trace{
            \mtx Y^{*, \intercal} \psione^* \kpsiinv \psione^\intercal \mtx Y
        }
        - \frac{\beta d_y}{2} \left( \psi_0^* - \trace{\kuu^{-1} \psitwo^*} \right),
    \end{aligned}
    \label{eqn:apx:SGPLVM:BoundTestTerm4}
\end{align}
where $\mtx D^{-1} = \mtx D^{-1/2} \mtx D^{-1/2}$ is easily computed since $\mtx D$ is diagonal.
Note also that many terms in Eq.~(\ref{eqn:apx:SGPLVM:BoundTestTerm3}) do not depend on the test case and can be cached after training is complete to improve efficiency.

\section*{References}

\end{document}